\documentclass[10pt,journal,compsoc,table,tikz,x11names,final]{IEEEtran}

\usepackage[pdftex]{graphicx}
\graphicspath{{.}}
\usepackage{pgf}  
\usepackage{epsfig}

\usepackage{amsfonts}
\usepackage{amsmath}
\usepackage{amssymb}
\usepackage{numprint}  
\usepackage[symbol]{footmisc}  
\npthousandsep{,}  
\usepackage{soul}  

\usepackage{makecell}  
\usepackage{multirow}  

\usepackage[caption=false]{subfig}

\usepackage{booktabs}
\usepackage[pagebackref=true,breaklinks=true,colorlinks,bookmarks=false]{hyperref}

\usepackage[nocompress]{cite}
\usepackage[numbers]{natbib}

\usepackage{forest}  
\usetikzlibrary{arrows.meta,angles,quotes}
\colorlet{linecol}{black!75}

\definecolor{lightergray}{rgb}{0.90, 0.90, 0.90}

\tikzset{
  my rounded corners/.append style={rounded corners=2pt},
}

\begin{document}
    \title{Deep Learning Methods for Abstract Visual Reasoning: A Survey on Raven's Progressive Matrices}
    \author{
        Miko{\l}aj~Ma{\l}ki{\'n}ski and Jacek~Ma{\'n}dziuk%
        \IEEEcompsocitemizethanks{%
            \IEEEcompsocthanksitem Miko{\l}aj~Ma{\l}ki{\'n}ski is a Ph.D. student at the Doctoral School no. 3, Warsaw University of Technology, pl. Politechniki 1, 00-661 Warsaw, Poland, m.malkinski@mini.pw.edu.pl.%
            \IEEEcompsocthanksitem Jacek~Ma{\'n}dziuk is with the Faculty of Mathematics and Information Science, Warsaw University of Technology, Koszykowa 75, 00-662 Warsaw, Poland, mandziuk@mini.pw.edu.pl.%
        }%
    }

    \IEEEtitleabstractindextext{
        \begin{abstract}
    Abstract visual reasoning (AVR) domain encompasses problems solving which requires the ability to reason about relations among entities present in a given scene.
    While humans, generally, solve AVR tasks in a ``natural'' way, even without prior experience, this type of problems has proven difficult for current machine learning systems.
    The paper summarises recent progress in applying deep learning methods to solving AVR problems, as a proxy for studying machine intelligence.
    We focus on the most common type of AVR tasks---the Raven's Progressive Matrices (RPMs)---and provide a comprehensive review of the learning methods and deep neural models applied to solve RPMs, as well as, the RPM benchmark sets.
    Performance analysis of the state-of-the-art approaches to solving RPMs leads to formulation of certain insights and remarks on the current and future trends in this area.
    We conclude the paper by demonstrating how real-world problems can benefit from the discoveries of RPM studies.
\end{abstract}

    }
    \maketitle
    
    \IEEEraisesectionheading{\section{Introduction}\label{sec:introduction}}

\IEEEPARstart{A}{long-standing} goal of human research endeavours is to understand the nature of intelligence.
Even though existing literature points to a variety of definitions~\cite{detterman1986intelligence,legg2007collection,hernandez2017measure}, most related to this work is intelligence as portrayed by the ability of applying existing knowledge, skills, and past experiences in entirely new settings.
This perspective was taken in a number of cognitive studies that measured intelligence (IQ) with the help of abstract visual reasoning (AVR) tasks~\cite{snow1984topography,carpenter1990one}.
AVR problems consist of images with simple 2D shapes governed by underlying abstract rules.
In order to solve them, the test-taker has to identify and often understand never-encountered abstract patterns and generalise them to new settings.

\begin{figure}[t]
    \centering
    \includegraphics[width=.45\textwidth]{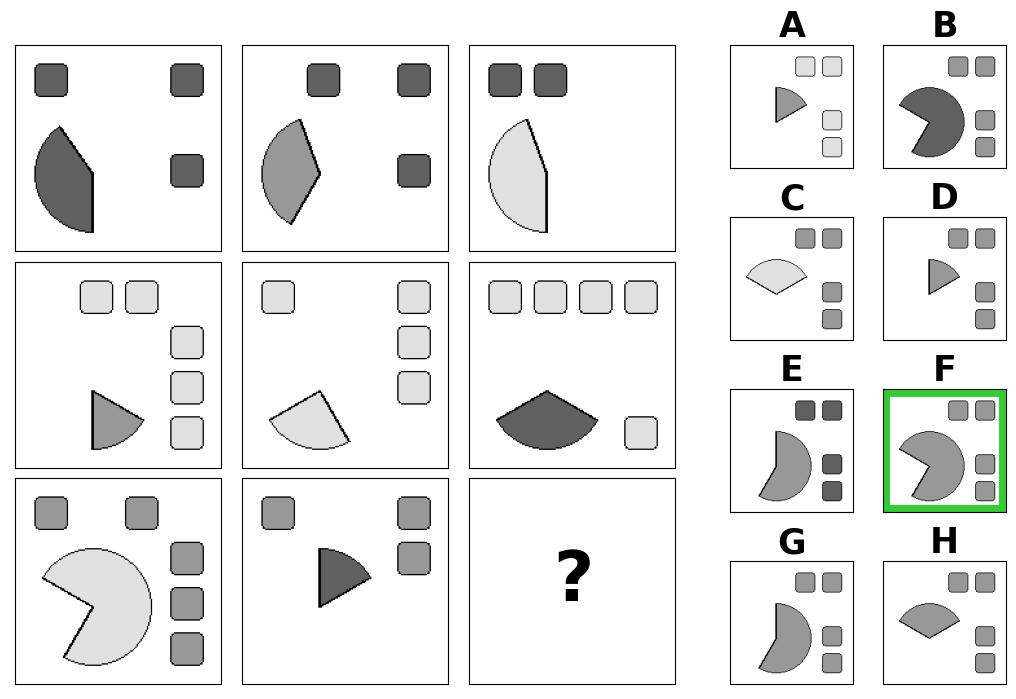}
    \caption{\textbf{RPM example.}
    Remarkably, humans are able to intuitively solve the challenge even without the exact definition of the task (that is presented in Section~\ref{sec:rpms}).
    The matrix is governed by multiple abstract patterns.
    Each row contains circle slices of 3 different colours split among columns, whereas squares have constant colour in each row.
    Moreover, positions of objects in the third column are determined by logical XOR applied row-wise in the case of squares and logical OR in the case of circle slices. The correct answer is~F.}
    \label{fig:rpm-example}
\end{figure}

Although there exists a wide range of AVR tasks~\cite{hofstadter1979godel}, in the cognitive literature one of them was studied with particular attention---the Raven's Progressive Matrices (RPMs)~\cite{raven1936mental,raven1998raven} (see Fig.~\ref{fig:rpm-example}).
RPMs were found to be highly diagnostic of abstract and relational reasoning abilities~\cite{snow1984topography} and representative for human intelligence in general~\cite{carpenter1990one}.
In light of these observations, recent works have started to investigate whether automatic pattern discovery algorithms are capable of achieving performance comparable to humans in solving RPMs.

A recent stream of research devoted to developing intelligent pattern analysis methods employs deep learning (DL)~\cite{lecun2015deep} for discovering regularities in complex settings.
Motivated by the impressive performance of DL methods in various domains~\cite{raghu2020survey} a question whether DL approaches could be effectively applied to solving RPMs has been posed~\cite{hoshen2017iq,barrett2018measuring}.
Since these seminal works, a number of approaches have been proposed which are reviewed and compared in this survey.

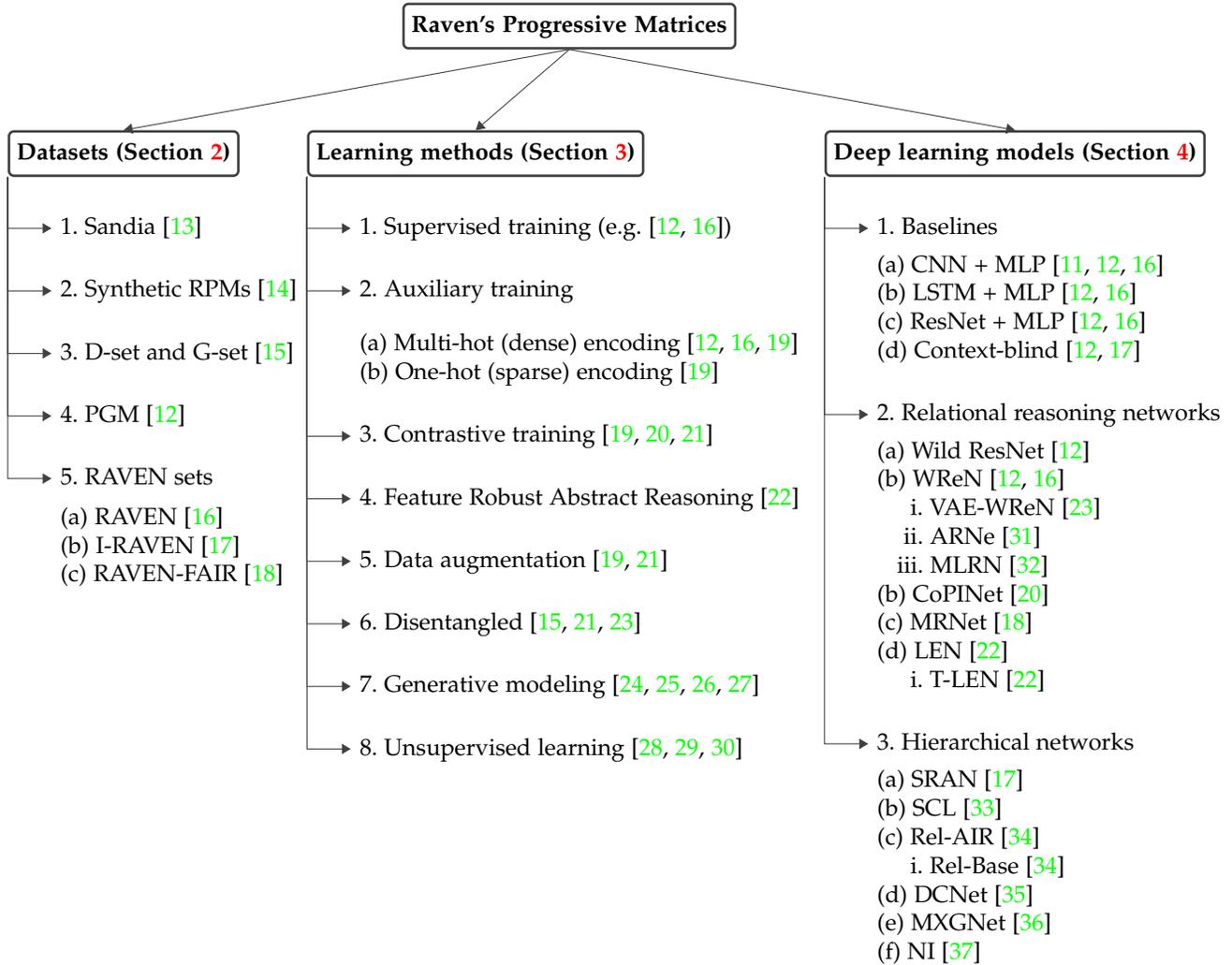
\begin{figure*}
    \centering
    \begin{forest}
        for tree={
            line width=1pt,
            if={level()<2}{
                my rounded corners,
                draw=linecol,
            }{},
            edge={color=linecol, >={Triangle[]}, ->},
            if level=0{%
                l sep+=.75cm,
                align=center,
                font=\bfseries,
                parent anchor=south,
            }{%
                if level=1{%
                    parent anchor=south west,
                    child anchor=north,
                    tier=parting ways,
                    align=center,
                    font=\bfseries,
                    for descendants={
                        child anchor=west,
                        parent anchor=west,
                        anchor=west,
                        align=left,
                    },
                }{
                    if level=2{
                        shape=coordinate,
                        no edge,
                        grow'=0,
                        calign with current edge,
                        xshift=10pt,
                        for descendants={
                            parent anchor=south west,
                            l sep+=-15pt
                        },
                        for children={
                            edge path={
                                \noexpand\path[\forestoption{edge}] (!to tier=parting ways.parent anchor) |- (.child anchor)\forestoption{edge label};
                            },
                            for descendants={
                                no edge,
                            },
                        },
                    }{},
                },
            }%
        },
  [Raven's Progressive Matrices
                                [Datasets (Section~\ref{sec:rpms})
                                [
                                [1. Sandia~\cite{matzen2010recreating}]
                                [2. Synthetic RPMs~\cite{wang2015automatic}]
                                [3. D-set and G-set~\cite{mandziuk2019deepiq}]
                                [4. PGM~\cite{barrett2018measuring}]
                                [5. RAVEN sets
                                [(a) RAVEN~\cite{zhang2019raven}\\ 
                                 (b) I-RAVEN~\cite{hu2021stratified} \\ 
                                 (c) RAVEN-FAIR~\cite{benny2020scale}]
                                ]
                                ]
                                ]
                                [Learning methods (Section~\ref{sec:learning-to-solve-rpms})
                                [
                                [1. Supervised training (e.g.~\cite{barrett2018measuring,zhang2019raven})]
                                [2. Auxiliary training
                                [(a) Multi-hot (dense) encoding~\cite{barrett2018measuring,zhang2019raven,malkinski2020multilabel} \\ (b) One-hot (sparse) encoding~\cite{malkinski2020multilabel}]
                                ]
                                [3. Contrastive training~\cite{zhang2019learning,malkinski2020multilabel,kim2020few}]
                                [4. Feature Robust Abstract Reasoning~\cite{zheng2019abstract}]
                                [5. Data augmentation~\cite{malkinski2020multilabel,kim2020few}]
                                [6. Disentangled~\cite{steenbrugge2018improving,mandziuk2019deepiq,kim2020few}]
                                [7. Generative modeling~\cite{pekar2020generating,hua2020modeling,shi2021raven,zhang2021abstract}]
                                [8. Unsupervised learning~\cite{zhuo2020solving,kiat2020pairwise,zhuo2021unsupervised}]
                                ]
                                ]
                                [Deep learning models (Section~\ref{sec:models})
                                [
                                [1. Baselines
                                [(a) CNN + MLP~\cite{hoshen2017iq,barrett2018measuring,zhang2019raven} \\ (b) LSTM + MLP~\cite{barrett2018measuring,zhang2019raven} \\ (c) ResNet + MLP~\cite{barrett2018measuring,zhang2019raven} \\ (d) Context-blind~\cite{barrett2018measuring,hu2021stratified}]
                                ]
                                [2. Relational reasoning networks
                                [(a) Wild ResNet~\cite{barrett2018measuring} \\ (b) WReN~\cite{barrett2018measuring,zhang2019raven} \\ ~~~~~i. VAE-WReN~\cite{steenbrugge2018improving} \\ ~~~~ii. ARNe~\cite{hahne2019attention} \\ ~~~iii. MLRN~\cite{jahrens2020solving} \\ (b) CoPINet~\cite{zhang2019learning} \\ (c) MRNet~\cite{benny2020scale} \\ (d) LEN~\cite{zheng2019abstract} \\ ~~~~~i. T-LEN~\cite{zheng2019abstract}]
                                ]
                                [3. Hierarchical networks
                                [(a) SRAN~\cite{hu2021stratified} \\ (b) SCL~\cite{wu2020scattering} \\ (c) Rel-AIR~\cite{spratley2020closer} \\ ~~~~~i. Rel-Base~\cite{spratley2020closer} \\ (d) DCNet~\cite{zhuo2021effective} \\ (e) MXGNet~\cite{wang2020abstract} \\ (f) NI~\cite{rahaman2021dynamic}]
                                ]
                                ]
                                ]
                                ]
    \end{forest}
    \caption{\textbf{RPM taxonomy.} A list of RPM benchmarks, learning methods and DL models considered in this paper.
    }
    \label{fig:rpm-taxonomy}
\end{figure*}

\subsection{Motivation and scope}
A growing number of recent publications use RPMs as a proxy to study machine intelligence.
Due to the increased interest in these problems among the DL community since the above-cited seminal papers~\cite{hoshen2017iq,barrett2018measuring}, a number of approaches have been proposed that vary in multiple aspects.
Furthermore, several RPM benchmarks with particular characteristics have been recently proposed that allow to analyse various properties and capabilities of the tested methods.
These methods vary in both the learning setup to solve the tasks and the model architecture.
The latter oftentimes differs from the architectures used in other domains that do not require reasoning about abstract relations spanning multiple entities.
This RPM-specific aspect imposes the use of dedicated neural components and makes analysis of RPM model architectures particularly interesting.

This survey collates existing works on RPMs -- the prevalent AVR benchmark in DL literature. The review is performed along all three above-mentioned angles, i.e. benchmark datasets, learning methods, and DL reasoning models, outlined in Fig.~\ref{fig:rpm-taxonomy}.

Some works that focus on RPMs have already reported superhuman performance on simpler benchmarks~\cite{mandziuk2019deepiq}, whereas the essence of AVR, i.e.\ the ability to generalise to novel difficult environments, remains unattained~\cite{barrett2018measuring}.
Indeed, a recent work~\cite{mitchell2021abstraction} shows that current AI systems (including DL methods, symbolic approaches and probabilistic program induction) still lag far behind the human capabilities not only in solving AVR problems, but generally, in broader settings of forming abstractions and analogies.
In order to assess and better understand the ``degree'' of intelligence represented by the current DL approaches for solving RPMs, we analyse and discuss in detail numerical results reported in the literature.

\subsection{Related work}

Initial attempts that tackled RPMs relied on manually prepared rules and heuristics~\cite{evans1964heuristic,foundalis2006phaeaco}, in some cases identified by analysing approaches of well-performing human solvers~\cite{strannegaard2013anthropomorphic}.
Another stream of methods that build on the insights from cognitive studies utilised the structure mapping theory~\cite{gentner1980structure,falkenhainer1986structure} to propose a set of RPM solvers that automatically discover relevant rules~\cite{lovett2007analogy,lovett2010structure}.
In addition, another set of methods eliminates the need for structure mapping and instead focuses on visual similarity of the RPM elements after inducing various image transformations~\cite{kunda2010taking,kunda2012reasoning}, potentially with different image resolutions that facilitates a fractal representation of the RPMs~\cite{mcgreggor2010fractal,mcgreggor2014confident}.
The progress made by these seminal works is comprehensively summarised in~\cite{hernandez2016computer}.

Despite the existence of many creative approaches for solving RPMs, in recent days DL methods constantly prove their ability to tackle some of the most challenging RPM benchmarks, surpassing human performance in some problem settings.
The recent abundance and superiority of DL methods motivated us to focus this survey exclusively on DL approaches. To our knowledge, such a review perspective has not been considered previously in the RPM literature.

While particular attention in DL literature on AVR problems is devoted to solving RPMs, the whole AVR domain is nowhere near limited to this task.
In fact, recent works introduce a broad spectrum of complementary abstract visual reasoning tasks that allow to test different characteristics of DL approaches.

Similarly to RPMs, the need for identifying abstract patterns is a recurring theme in all AVR tasks.
However, depending on the specific problem, these patterns have to be extracted from different configurations of matrix panels and then applied in various contexts.
In the odd one out tasks~\cite{gardner2006colossal,ruiz2011building,smets2011odd}, the solver has to identify an odd element that breaks a rule instantiated in the remaining images.
Similar idea is presented in Bongard Problems~\cite{bongard1968recognition} where the goal is to describe a rule that is instantiated in a set of images and broken in a supplementary set of panels.
Same-different tasks~\cite{fleuret2011comparing} present a related challenge, in which each problem instance contains two sets of images separated by an abstract pattern.
Given a new image, the test-taker has to assign it to one of the two presented sets.
Visual analogy problems~\cite{hill2019learning} are structurally similar to RPMs and alike test the ability of making analogies based on abstract patterns that govern the objects and their attributes in images.
However, in contrast to RPMs, these problems allow to test the ability of generalising an abstract concept from a given source domain to a different target domain.
Additional problems test the extrapolation ability~\cite{webb2020learning}, capacity of recognizing abstract patterns from only few samples~\cite{chollet2019measure}, or introduce numbers into the matrices and test the combined ability of abstract and numerical visual reasoning~\cite{zhang2020machine}.
A comprehensive review of these emerging AVR tasks is conducted in~\cite{malkinski2022review}.

Although each of the above-mentioned problems has its own specificity, we believe that in-depth overview of DL application to solving RPMs---a predominant AVR challenge---will inspire the readers to identify promising paths for solving related AVR tasks.

\subsection{Structure}
The rest of this work is structured as follows.
In Section~\ref{sec:rpms} we introduce the Raven's Progressive Matrices, discuss their importance in measuring intelligence and describe benchmarks together with their automatic generation methods.
In Sections~\ref{sec:learning-to-solve-rpms} and~\ref{sec:models} we characterise a variety of learning setups to solve RPMs and the related DL architectures, respectively.
Current evaluations of machine intelligence are aggregated and analysed in Section~\ref{sec:results}.
Section~\ref{sec:discussion} links the RPM-related literature with advances in other fields, presents open questions, potential practical applications of AVR research, and directions for future studies.
The survey is concluded in Section~\ref{sec:conclusion}.

    \section{Raven's Progressive Matrices}\label{sec:rpms}

The ability to solve Raven's Progressive Matrices~\cite{raven1936mental,raven1998raven} is believed to be highly correlated with human intelligence~\cite{carpenter1990one} and is therefore also considered as a natural measure of intelligence of advanced artificial reasoning systems.
RPMs allow to measure both structural and abstract reasoning skills~\cite{snow1984topography}, which characterise human fluid / high-level intelligence~\cite{jaeggi2008improving}.

\subsection{Problem statement}
\begin{figure}[t]
    \centering
    \includegraphics[width=.5\textwidth]{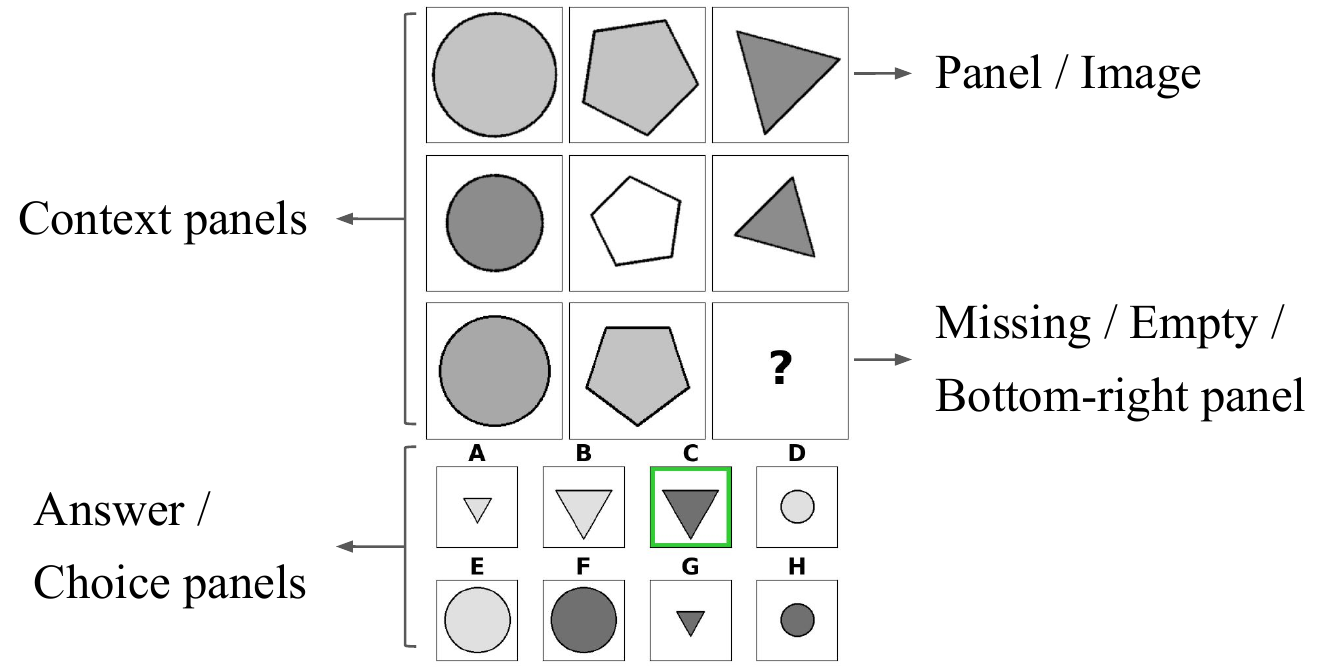}
    \caption{\textbf{RPM notation.} A sample RPM from the RAVEN dataset~\cite{zhang2019raven}, which consists of two parts -- the context and the answer panels.
    The goal is to complete the context with appropriate answer panel.
    The chosen answer panel must fulfil all abstract rules governing the content of the context panels.
    In the example there are three such rules applied: (1) the same set of 3 shapes is present in each row, (2) the shapes in a given row are of the same size, and (3) the shape color intensity in the third column equals the sum of color intensities in two previous columns.
    Hence, the correct answer is C.
    }
    \label{fig:rpm-notation}
\end{figure}

Although RPMs can have variable structure, the most common matrices investigated in recent machine learning (ML) literature are composed of images arranged into two distinct parts.
The first part of the matrix usually contains 8 images arranged in a $3 \times 3$ grid, referred to as the context panels, where the bottom-right image is missing.
The goal is to select a panel which correctly completes this matrix from another set of several images, referred to as the answer panels (see Fig.~\ref{fig:rpm-notation}).
In order to find the correct answer the test-taker is required to identify a set of underlying abstract rules which govern the visual attributes of the matrix.
Generally, these rules describe how image features differ between the matrix panels.
These relations, as well as objects and their attributes are defined differently depending on the dataset.

\subsection{Automatic generation of RPMs}
The original set of RPMs (called the Standard Progressive Matrices, SPMs) proposed in~\cite{raven1936mental,raven1998raven} contains a limited number of hand-crafted instances, which may be insufficient for training even simple statistical pattern recognition models that struggle to learn from small sample size~\cite{jain198239,raudys1991small}.
Since the process of successful training of DL models usually requires large volumes of training data points~\cite{sun2017revisiting,mahajan2018exploring}, there is a strong need for methods of automatic generation of AVR problems that could provide sufficiently large numbers of training samples.

In order to procedurally generate new RPMs, several challenges have to be addressed.
First, the generated problems should be visually diverse to make them non-repetitive and appealing to the test-taker.
Moreover, multiple levels of difficulty are preferred that allow to better estimate the reasoning capabilities of the solver.
Lastly, the generated RPM should be valid, i.e. there should be only one answer from the set of possible choices that correctly completes the matrix.
The task of choosing the correct answer should be realisable by following the Occam's razor principle -- the answer should be justifiable with a minimal set of rules that govern relations between objects and their attributes across context panels.

\begin{figure*}[t]
    \centering
    \subfloat[Sandia~\cite{matzen2010recreating}]{\includegraphics[width=.18\textwidth]{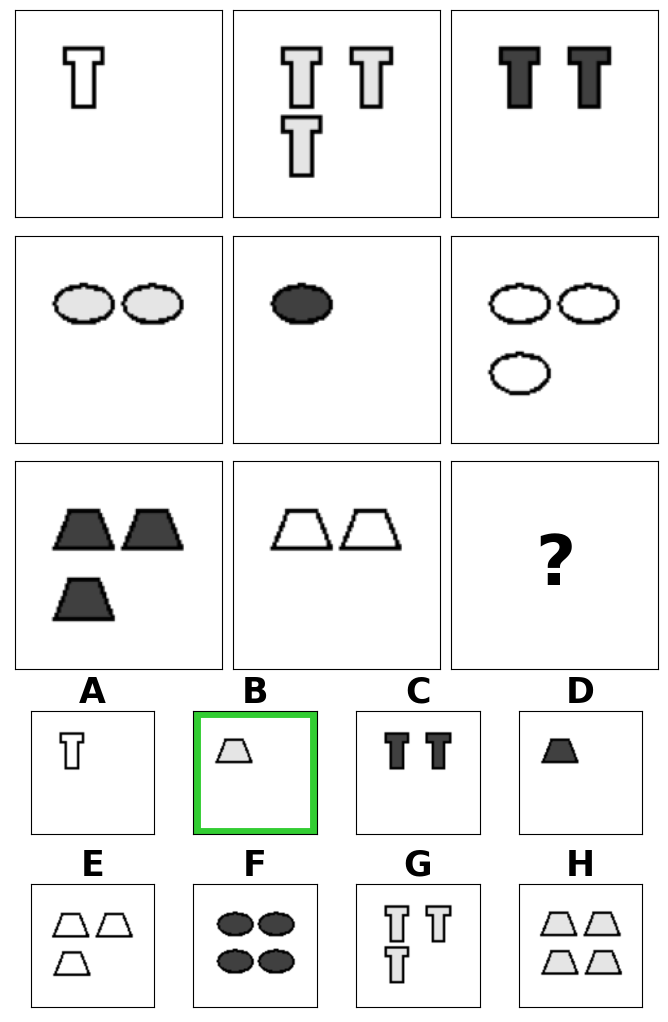}\label{fig:rpm-sandia}}
    \hfil
    \subfloat[Synthetic~\cite{wang2015automatic}]{\includegraphics[width=.18\textwidth]{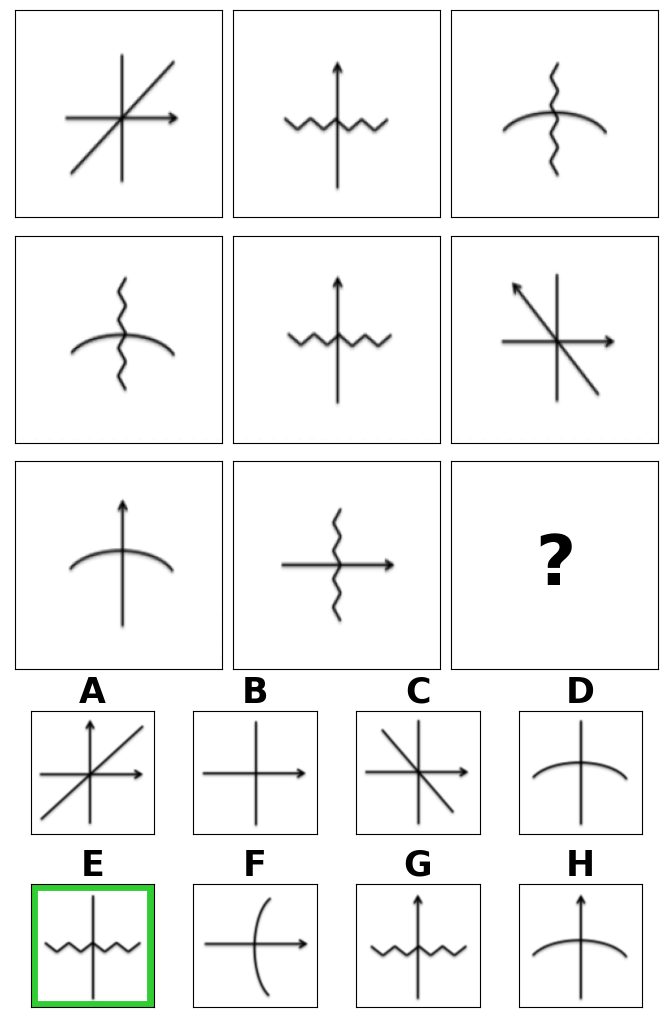}\label{fig:rpm-wang2015automatic}}
    \hfil
    \subfloat[G-set~\cite{mandziuk2019deepiq}]{\includegraphics[width=.18\textwidth]{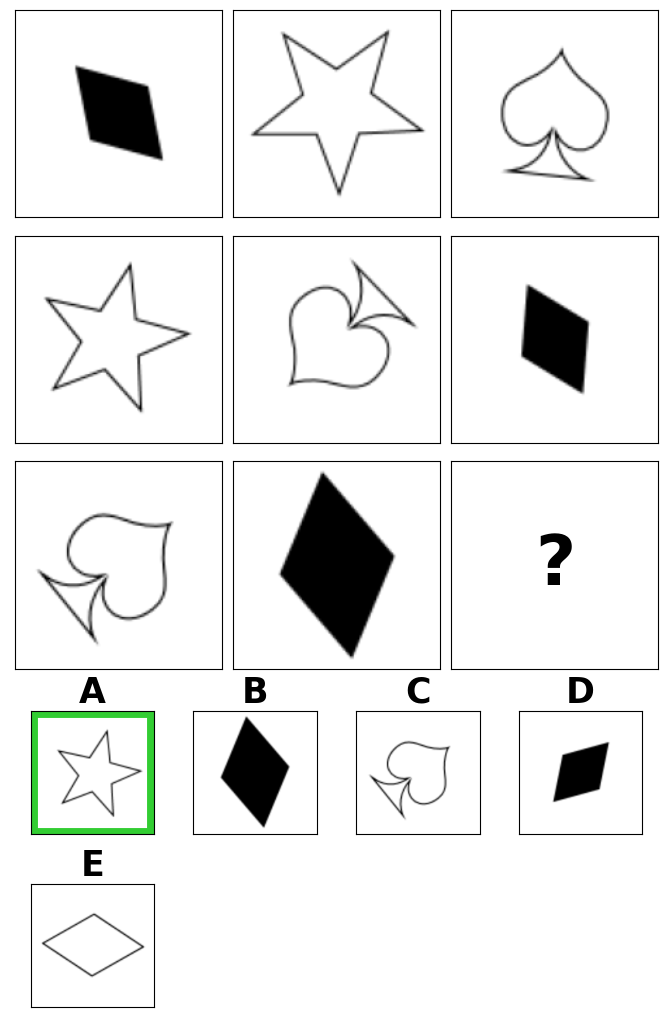}\label{fig:rpm-deepiq}}
    \hfil
    \subfloat[PGM~\cite{barrett2018measuring}]{\includegraphics[width=.18\textwidth]{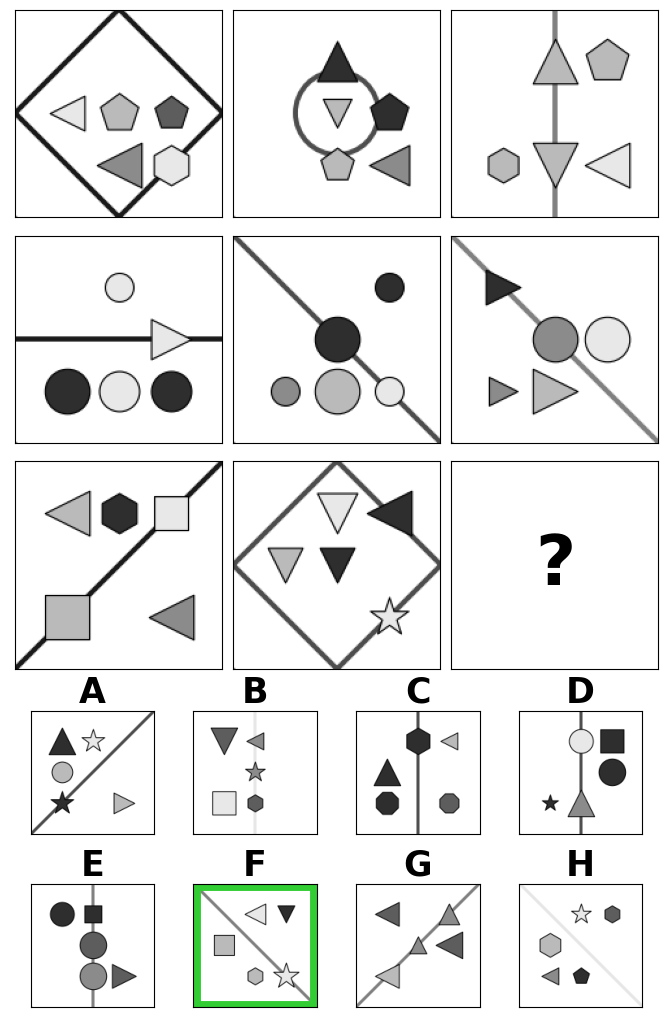}\label{fig:rpm-pgm}}
    \hfil
    \subfloat[I-RAVEN~\cite{zhang2019raven,hu2021stratified}]{\includegraphics[width=.18\textwidth]{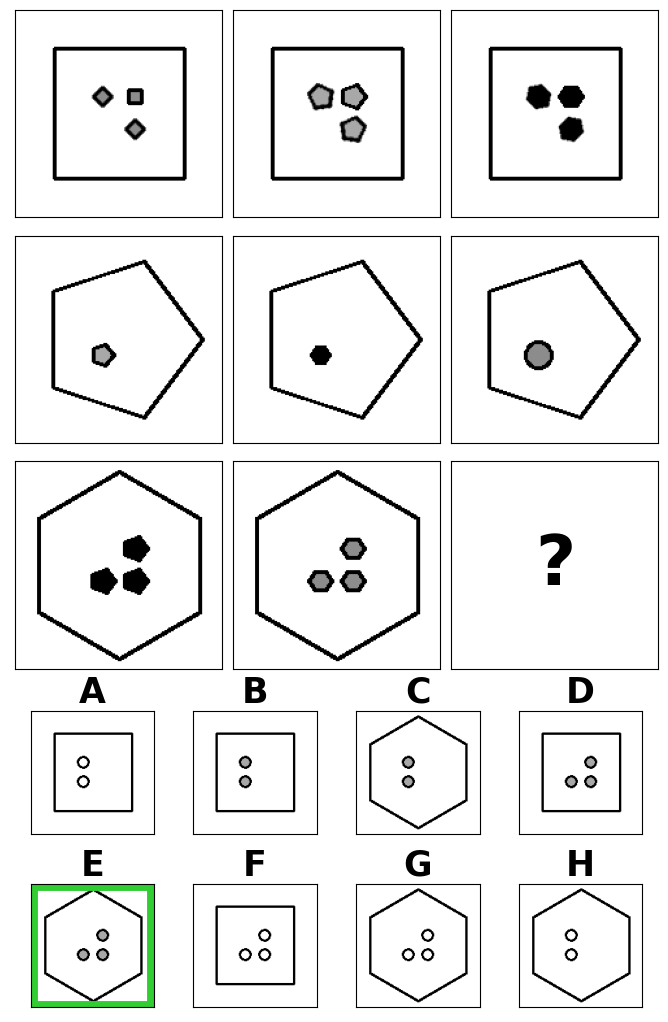}\label{fig:rpm-i-raven}}
    \caption{\textbf{RPM examples.} Correct answers are marked with a green boundary. Matrices from RAVEN, I-RAVEN and RAVEN-FAIR differ only in the way of generating answers, hence only a selected matrix from the I-RAVEN dataset is shown.}
    \label{fig:rpm-examples}
\end{figure*}

Geared toward satisfying these requirements, various attempts have been made to design an RPM generation algorithm capable of delivering huge number of instances.
In effect, several popular RPM datasets have been proposed which include the Sandia matrices~\cite[]{matzen2010recreating}, D-set and G-set from~\cite{mandziuk2019deepiq}, PGM~\cite{barrett2018measuring}, RAVEN~\cite{zhang2019raven}, I-RAVEN~\cite{hu2021stratified} and RAVEN-FAIR~\cite{benny2020scale}.
Samples from these datasets are illustrated in Fig.~\ref{fig:rpm-examples}.

\subsubsection{Sandia matrices}
The Sandia matrix generation process~\cite{matzen2010recreating} marked the first widely-known attempt to expand the set of available RPM instances.
Based on the analysis of the SPMs, the authors extracted a set of logic rules, shapes, and transformations that modify their attributes, which were later used to generate a large set of RPMs.
The authors compared the quality of generated matrices to the original ones and found out, that although simpler instances (with 1 or 2 rules) were of similar difficulty to SPMs, instances with higher number of rules were generally more difficult than their SPMs counterparts.
A sample RPM generated with the Sandia software is shown in Fig.~\ref{fig:rpm-sandia}.

\subsubsection{Synthetic RPMs}
Another approach represents abstract RPM structure using first-order logic and formulates sampling restrictions that allow to construct only valid RPMs~\cite{wang2015automatic}, such as those presented in Fig.~\ref{fig:rpm-wang2015automatic}.
Moreover, the authors conducted a user study to validate whether the constructed matrices differ from the original set of SPMs proposed in~\cite{raven1936mental}.
The experiment revealed that the generated problems are statistically indistinguishable from the manually designed matrices among a group of 24 respondents.

\subsubsection{D-set and G-set}
Another suite of automatically generated RPMs was proposed in~\cite{mandziuk2019deepiq}.
The D-set and G-set datasets have been generated based on design principles from prior works~\cite{ragni2014analyzing,carpenter1990one}.
The resultant matrices, with an example shown in Fig.~\ref{fig:rpm-deepiq}, although structurally similar to those from the Sandia suite, represented a different feature distribution.
Namely, the authors of~\cite{mandziuk2019deepiq} utilised a completely different set of object shapes and supported full ranges for object attributes (shading, rotation and size) as compared to discrete values used in Sandia.
This broader RPM configuration was realised in D-set, whereas G-set was curated to resemble the feature distributions found in Sandia.
Thanks to synthesizing their own RPMs, the authors were able to use Sandia matrices as an additional evaluation dataset for verifying out-of-distribution generalisation of the proposed DeepIQ system in a transductive transfer learning setting.

\subsubsection{PGM}
Machine performance on the above-introduced datasets~\cite{mandziuk2019deepiq,matzen2010recreating} may be misleading, as the models are trained on large sets of matrices with similar structure to those found in the testing suite.
In order to better evaluate the generalisation capabilities of DL models the PGM dataset~\cite{barrett2018measuring} was proposed which arranges problems into 8 generalisation regimes with variable difficulty.
To achieve this goal, each matrix from PGM (see example in Fig.~\ref{fig:rpm-pgm}) has an associated representation of the abstract structure, i.e. a set of triples $\mathcal{S} = \{[r, o, a]\ \vert\ r\in\mathcal{R}, 0\in\mathcal{O}, a\in\mathcal{A}\}$, where $\mathcal{R}$ $=$ $\{\texttt{progression},$ $\texttt{XOR},$ $\texttt{OR},$ $\texttt{AND},$ $\texttt{consistent union}\}$ defines the set of rules, $\mathcal{O}$ $=$ $\{\texttt{shape},$ $\texttt{line}\}$ the set of objects and $\mathcal{A}$ $=$ $\{\texttt{size},$ $\texttt{type},$ $\texttt{color},$ $\texttt{position},$ $\texttt{number}\}$ the set of attributes.
Based on these structures the dataset arranges RPM instances intro training, validation and test splits, such that both train and validation parts have matrices with the same structures, whereas test matrices are governed by relations not seen in the train and validation sets.
Although being a perfect test bed for measuring generalisation in DL models, the dataset is characterised with an enormous size (each regime contains $1\,420\,000$ RPMs, where $1.2$M belong to the training split) that is often a bottleneck for evaluating multiple models.

\subsubsection{RAVEN}
Another RPM dataset named RAVEN~\cite{zhang2019raven} aims to present a visually broader matrices with hierarchical structure (see Fig.~\ref{fig:rpm-i-raven}). This is achieved with the help of Attributed Stochastic Image Grammar (A-SIG)~\cite{fu1974syntactic,zhu2007stochastic,lin2009stochastic}.
RAVEN contains matrices belonging to 7 visual configurations and similarly to PGM each RPM from RAVEN has an associatied abstract structure.
However, in this case the structure is defined as a set of pairs $\mathcal{S}$ $=$ $\{[r, a]$ $\vert$ $r \in \mathcal{R},$ $a \in \mathcal{A}\}$, where $\mathcal{R}$ $=$ $\{\texttt{constant},$ $\texttt{progression},$ $\texttt{arithmetic},$ $\texttt{distribute three}\}$ defines the set of rules and $\mathcal{A}$ $=$ $\{\texttt{number},$ $\texttt{position},$ $\texttt{type},$ $\texttt{size},$ $\texttt{color}\}$ the set of attributes.
In contrast to PGM, the dataset contains supplementary structural annotations thanks to A-SIG that connects visual and structural representations.
Moreover, \citet{zhang2019raven} have carried out a human evaluation on RAVEN matrices which allows to better assess the performance of DL approaches.

The dataset is composed of $42\,000$ training RPMs, and additional $2 \times 14\,000$ problems allocated for validation and testing splits, respectively. In comparison to PGM, RAVEN's matrices have a few times higher average number of rules: 6.29 vs 1.37.
At the same time, PGM is better-suited for evaluating out-of-distribution generalisation due to defining explicit generalisation regimes.

\subsubsection{I-RAVEN}
Although a follow-up work reported superhuman performance on RAVEN~\cite{zhang2019learning}, it was later revealed that such impressive results may arise from a shortcut solution due to biased answer sets.
In fact, the problem of shortcut learning is prevalent in visual reasoning research and was identified across multiple related problems~\cite{agrawal2018don,d2020underspecification,geirhos2020shortcut,dancette2021beyond}.
In the case of RAVEN, a context-blind model---one which processes only the answer panels and discards context ones---was shown to achieve close to perfect performance, bypassing the need for discovering abstract rules that govern the matrices.
It was brought to light that correct answer to RPMs from RAVEN may be obtained by selecting answer panel with the most common attributes~\cite{hu2021stratified}.
In order to fix this defect, the I-RAVEN dataset~\cite{hu2021stratified} was proposed that generates the set of answers with an iterative tree-based method.
Context-blind models trained on such impartial dataset were shown to produce classifications equivalent to random guessing~\cite{wu2020scattering,hu2021stratified}, in effect demonstrating validity of this dataset.

\subsubsection{RAVEN-FAIR}
Similarly to I-RAVEN, RAVEN-FAIR~\cite{benny2020scale} promises to solve the problem of biased choice panels of the original dataset.
However, as noted in the supplementary material of~\cite[Table~4]{benny2020scale}, a Context-blind ResNet model scores 17.2\% accuracy on the proposed dataset.
At the same time, the same context-blind model evaluated on I-RAVEN scores 12.5\%, which is equal to the random guess accuracy.
This indicates that although RAVEN-FAIR is unquestionably less biased than the original RAVEN, not all bias sources were mitigated.
Therefore, among three RAVEN-type datasets, it is recommended to use the unbiased I-RAVEN when testing RPM reasoning models.

    \section{Learning to solve RPMs}\label{sec:learning-to-solve-rpms}

After the preliminary attempts~\cite{hoshen2017iq,barrett2018measuring}, multiple distinct approaches have been proposed for training DL models to solve RPMs, which are summarised in this section.
Assume that an RPM instance $\mathcal{P} = (X, y)$ has $16$ panels divided equally into context and answer images, where $X = \{x_i\}_{i=1}^{16}$ is a set of all $16$ images and $y \in \{1, \ldots, 8\}$ is an index of the correct answer.
Moreover, we denote by $X_c = \{x_i\}_{i=1}^8 \subset X$ the set of context panels, and by $X_a = \{x_i\}_{i=9}^{16} = \{a_j\}_{j=1}^{8} \subset X$ the set of answer panels (a.k.a candidate or choice panels).
The above number of $8$ answer panels holds for RPMs from Sandia suite~\cite{matzen2010recreating}, for synthetic RPMs~\cite{wang2015automatic}, for matrices from PGM~\cite{barrett2018measuring}, RAVEN~\cite{zhang2019raven}, I-RAVEN~\cite{hu2021stratified}, and RAVEN-FAIR~\cite{benny2020scale}, whereas RPMs used in~\cite{mandziuk2019deepiq} have 3 possible answers less.
Let us also denote by $X_{c \cup a_j} = \{x_i\}_{i=1}^{8} \cup \{a_j\} \subset X$ an RPM with the missing panel completed by an answer panel with index $j$.

Consider an RPM reasoning model $\mathcal{N}(X) = \{h_j\}_{j=1}^{8}$, that given the RPM panels $X$ produces embedding vectors $\{h_j\}_{j=1}^{8}$, one for each of the answer panels, where $h_j \in \mathbb{R}^{d}$ for $d \geq 1$.
Hereinafter we will refer to $\{h_j\}_{j=1}^{8}$ as \textit{candidate embeddings}.
Model $\mathcal{N}$ can be implemented as any differentiable function -- concrete examples from the literature are discussed in the following section.
Based on the provided definitions let us now focus on the proposed schemes for learning to solve RPMs.

\subsection{Supervised training}
In the supervised training, the model is trained to predict an index of the answer panel which correctly completes the matrix.
For this purpose, several works (e.g.~\cite{hoshen2017iq,barrett2018measuring,mandziuk2019deepiq,zhang2019raven}) employ a scoring module $\psi(h) = s \in \mathbb{R}$, which produces a single logit $s$ for each candidate embedding.
Although in practice $\psi$ is often implemented as a multi-layer network (e.g. MLP in~\cite{hoshen2017iq,mandziuk2019deepiq} or Relation Network~\cite{santoro2017simple} in~\cite{barrett2018measuring}), we consider these modules as part of $\mathcal{N}$ and consider the scoring module in the form of a simple linear layer with learnable weights.
The supervised setup gathers individual logits into a set $\mathcal{S} = \{s_j\}_{j=1}^8$ and converts it to a probability distribution over the set of possible answers for an RPM $\mathcal{P}$, with $p(\mathcal{P}) = \{p(\mathcal{P})_j\}_{j=1}^8 = \text{softmax}(\mathcal{S})$.
Using the estimated probability, the scoring module $\psi$ is optimised together with the base network $\mathcal{N}$ with a standard cross-entropy loss function.
That is, for a batch of RPMs $\{\mathcal{P}_i\}_{i=1}^N$, where $N$ is the batch size, the following objective is minimised:
\begin{equation}
    \mathcal{L}^{\text{ce}} = -\frac{1}{N}\sum_{i = 1}^N p(\mathcal{P}_i) \log q(\mathcal{P}_i)
\end{equation}
where $q(\mathcal{P}_i) = \text{onehot}(y_i)$ is the one-hot encoded index of the correct answer for $\mathcal{P}_i$.
The choice panel corresponding to the highest probability is considered as an answer chosen by the network, i.e. $\hat{y} = \text{argmax}_{j}\{p(\mathcal{P})_j\}_{j=1}^8$.

\subsection{Auxiliary training}
In order to solve an RPM, one has to first recognise its abstract structure which governs the objects, attributes and relations present in the images.
It was shown in~\cite{barrett2018measuring} that training neural networks to justify their answer by predicting such an abstract structure leads to better performance in the final classification task.
For this purpose \citet{barrett2018measuring} proposed to redefine RPMs as triples $\mathcal{P} = (X, R, y)$, where $R \subset \mathcal{R}$ defines the set of underlying abstract rules governing the RPM.
The set of all abstract rules $\mathcal{R}$ is defined dependent on the dataset, as well as the maximal number of rules $n_R$ per RPM ($1 \leq n_R \leq 4$ for PGM and $1 \leq n_R \leq 8$ for RAVEN and its derivatives).
\citet{barrett2018measuring} proposed to encode PGM abstract rules with a multi-hot encoding and employ a rule prediction head $\rho(\sum_{j=1}^8 h_j) = \widehat{R}$, for estimating the set of RPM abstract rules, in the form of a vector $\widehat{R}$ of fixed length.
The prediction $\widehat{R}$ was activated with a sigmoid function.
A binary cross-entropy loss function $\mathcal{L}^{\text{aux}}$ was used to compare it with the ground-truth encoded rule representation $R$.
During training a joint loss function $\mathcal{L} = \mathcal{L}^{\text{ce}} + \beta \mathcal{L}^{\text{aux}}$ was minimised, where $\beta$ was a balancing coefficient.

Similar approach was validated on the RAVEN dataset in~\cite{zhang2019raven}, where besides the rule related auxiliary target the authors proposed another loss function related to the prediction of RPM structure.
Surprisingly, after training models with these supplementary objectives, their performance deteriorated.
Analogous conclusions were drawn in multiple follow-up works~\cite{zhang2019learning,hu2021stratified,wu2020scattering,wang2020abstract} that evaluated their approaches on RAVEN and its derivatives.
However, in~\cite{malkinski2020multilabel} we have shown that this inferior performance can be overcome by replacing the rule encoding method with the sparse encoding---an alternative rule representation based on one-hot encoding---that provides a more accurate training signal.

\subsection{Contrastive training}
The ability to juxtapose correct and wrong answers was identified as a key component in adaptive problem solving across cognitive literature~\cite{gentner1983structure,hofstadter1995fluid,smith2014role}.
A number of works have proposed to incorporate such contrastive mechanisms either directly in the model architecture~\cite{zhang2019learning} or in the objective function~\cite{zhang2019learning,malkinski2020multilabel,kim2020few,hu2021stratified}.
\citet{zhang2019learning} formulate an alternative to the supervised training setup by substituting cross-entropy with a variant of NCE loss that encourages contrast effects.
The authors argue for shifting the view of solving RPMs from a classification task to ranking, where answer panels are ranked according to their probability of correctly completing the matrix.
NCE-based objective functions are further utilised in the following works.

\citet{hu2021stratified} employ a hierarchical model to generate embeddings of all possible pairs of RPM rows (and optionally columns).
In effect, an embedding of the first two RPM rows (a so-called dominant embedding) is obtained, as well as embeddings of pairs of rows containing the last row completed by one of the answer panels (let us call it the candidate row pair embeddings).
Then, an NCE-inspired $(N+1)$-tuplet loss~\cite{sohn2016improved} is used to maximise the similarity of the dominant embedding to the candidate row pair embeddings completed by the correct answer.
At the same time, the loss function minimises the similarity of the dominant embedding to the candidate row pair embeddings completed by each of the wrong answers.

Another work casts the problem of solving RPMs into a multi-label classification framework and proposes the Multi-Label Contrastive Loss~\cite{malkinski2020multilabel} -- a contrastive objective function which builds on the Supervised Contrastive Loss~\cite{khosla2020supervised}.
The authors propose a pre-training objective which builds similar representations to RPMs with common abstract structure and different representations for unrelated RPMs.
In contrast to~\cite{hu2021stratified}, the proposed method considers embeddings of the whole RPM context completed by an answer panel instead of the pairs of rows.
Moreover, the method is combined with auxiliary training with sparse encoding into a joint learning method called Multi-Label Contrastive Learning (MLCL).

Meta-Analogical Contrastive Learning~\cite{kim2020few}, similarly to MLCL, improves the efficacy of learning relational representations of AVR tasks by maximising similarity between analogical structural relations and minimising similarity between non-analogical ones.
The work defines multiple analogy types: 1) intra-problem analogy draws an analogy between the original RPM context panels and its perturbation obtained via replacing randomly selected panel with noise, 2) inter-problem analogy collates two RPM instances with the same abstract structures but different attribute values and 3) non-analogy randomly shuffles the RPM context panels in order to break the abstract structure.

\subsection{Learning with an optimal trajectory}
In~\cite{zheng2019abstract} the authors recognise distracting features---image objects whose attributes change at random---as the main challenge in learning to solve RPMs.
As a possible solution Feature Robust Abstract Reasoning (FRAR)~\cite{zheng2019abstract} is proposed, which mitigates the impact of distracting features by means of a carefully designed learning trajectory based on a student-teacher reinforcement learning approach.
Moreover, the paper conducts a wide study of optimal learning trajectory approaches including Curriculum learning~\cite{bengio2009curriculum}, Self-paced learning~\cite{kumar2010self}, Learning to teach~\cite{borko1996learning}, Hard example mining~\cite{malisiewicz2011ensemble}, Focal loss~\cite{lin2017focal} and Mentornet-PD~\cite{jiang2018mentornet}.
The experiments reveal that learning to solve RPMs is most effective with the proposed FRAR method.

\begin{figure}[t]
    \centering
    \includegraphics[width=0.45\columnwidth]{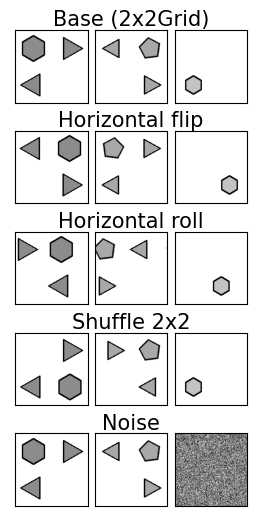}
    ~
    \includegraphics[width=0.45\columnwidth]{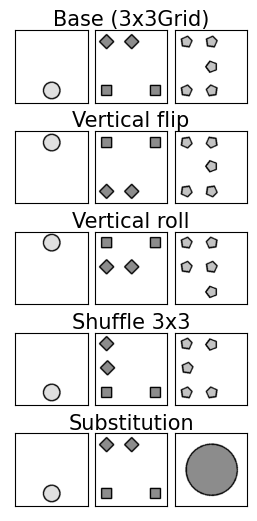}
    \caption{
    \textbf{RPM augmentation.}
    The first row presents single rows from two I-RAVEN matrices with configurations \texttt{2x2Grid} (left) and \texttt{3x3Grid} (right).
    Rows 2--4 demonstrate image-level augmentations used in~\cite{malkinski2020multilabel}, whereas the last row illustrates structural perturbations applied in~\cite{kim2020few}.
    }
    \label{fig:augmentation}
\end{figure}

\subsection{Data augmentation}
Data augmentation methods were shown to be of critical importance in applying DL across diverse domains~\cite{ko2015audio,shorten2019survey,wei2019eda,wen2020time}.
Similar conclusions were drawn in the field of AVR, where RPM data augmentation have proven to enhance abstract reasoning capabilities of neural learning models~\cite{malkinski2020multilabel,kim2020few}.
In~\cite{malkinski2020multilabel} it is proposed to use simple image transformations including horizontal/vertical flip, horizontal/vertical roll, shuffle 2x2/3x3, rotation and transposition, as illustrated in Fig.~\ref{fig:augmentation}, and apply them consistently to all panels of a given RPM\@.
The authors have shown that data augmentation boosts the RPM solving performance irrespective of the chosen model and training setup.
The topic of data augmentation was explored in parallel in~\cite{kim2020few}, where the authors propose to shuffle the context panels of an RPM in order to break its abstract structure and use such modified instances as negative pairs in contrastive learning or to replace selected RPM panels with noise.
These two approaches mainly differ in that~\cite{malkinski2020multilabel} relies on image-level transformations, whereas~\cite{kim2020few} modifies the RPM structure by rearranging or replacing its panels.

\subsection{Disentangled representations}
Recent studies have shown that disentangled representations are helpful for abstract reasoning tasks and can improve sample-efficiency~\cite{van2019disentangled}.
Specifically, it was demonstrated that a perception backbone of WReN~\cite{barrett2018measuring} pre-trained as a disentangled Variational Autoencoder ($\beta$-VAE)~\cite{kingma2013auto,higgins2016beta,burgess2018understanding} leads to better generalisation capabilities on the downstream task of solving RPMs than the same architecture trained in a purely supervised manner~\cite{steenbrugge2018improving}.
Moreover, autoencoders were successfully utilised in~\cite{mandziuk2019deepiq} for learning transferable features in simple AVR tasks, whereas the authors of~\cite{kim2020few} incorporated an autoencoder as part of their architecture for creating so-called generative analogies.

\subsection{Generative modeling}
Another work~\cite{pekar2020generating} employs variational autoencoders for the purpose of training an effective generative model~\cite{goodfellow2014generative,creswell2018generative,wang2021generative} capable of producing probable answer panels for RPMs.
The method combines multiple components responsible for: 1) reconstruction of an answer image with a variational autoencoder, 2) predicting an index of the correct answer (supervised training) and the representation of abstract rules (auxiliary training), and 3) generating a new possible answer based on the latent embedding from VAE and the latent embedding of the recognition pathway.
Such multi-task network was shown to be capable of generating plausible answer panels that preserve the underlying abstract rules and being competitive with models trained only with the supervised training.

An alternative RPM answer generation approach was proposed in~\cite{shi2021raven}, where a deep latent variable model that utilises multiple Gaussian processes was constructed.
The resultant method was shown to be interpretable via concept-specific latent variables and produced high-quality RPM panels.
While both works~\cite{pekar2020generating,shi2021raven} train generative models using RPM panels, it was also demonstrated that an image generator pre-trained on real-world images may as well be effective in producing valid RPM panels~\cite{hua2020modeling}.

Different from the above works, the Probabilistic Abduction and Execution (PrAE) model~\cite{zhang2021abstract} is able to construct a probabilistic RPM representation and use it to generate answers to RPMs with a scene inference engine.
In contrast to the previous end-to-end generative approaches, PrAE decouples the generation process into a neural perception backbone and a symbolic logical reasoning engine.
In spite of this separation, the method can be optimised end-to-end with REINFORCE~\cite{williams1992simple} and showcases the applicability of neuro-symbolic approaches to solving RPMs within a generative process.

\subsection{Unsupervised learning}
A couple of recent approaches to solving RPMs investigate whether useful representations can be learned in an usupervised manner.
In this setup the model does not take into account information about the correct answer in the training process.

Geared towards obtaining an effective RPM solving model in the unsupervised setting, Noisy Contrast and Decentralization (NCD)~\cite{zhuo2021unsupervised} considers 10 rows obtained from each RPM (first and second rows together with the third row completed by each of the 8 answer panels).
Next the method considers a binary classification task in which the model predicts which rows belong to RPM context and which are formed by completing the third row with one of the answers.
During inference, NCD generates probabilities only for the 8 versions of the last row and outputs an index of the highest scoring candidate as the answer.
The method builds upon Multi-label Classification with Pseudo Targets (MCPT)~\cite{zhuo2020solving} proposed by the authors of NCD in their earlier work.

Another unsupervised approach, the Pairwise Relations Discriminator (PRD), is proposed in~\cite{kiat2020pairwise}.
The method trains an underlying model to solve the task of discriminating between positive and negative pairs, similarly to other contrastive approaches (e.g.~\cite{hu2021stratified}).
Positive pairs are formed by taking the first two rows from a given RPM, say $\mathcal{P}$.
The negative pairs are obtained in several ways.
First option is to pair one of the the first two rows from $\mathcal{P}$ with another row from a different RPM.
Alternatively, the method randomly selects the first or the second row as $r$ and considers the other one as $r'$.
Then, the third panel of $r'$ is replaced by one of the answers to form a negative pair with $r$.
Analogously, the third row may be taken with its third panel replaced by the third image from $r'$ and considered a negative pair with $r$.
Next, PRD maximises the similarity of elements within positive pairs and minimises the similarity of items within the negative pairs.
To obtain an answer for $\mathcal{P}$, the method fills in the third row with each answer and calculates the average similarity between the completed row and the first two rows.
The choice panel corresponding to the completed row with the highest similarity is considered as the answer.

{\color{orange}}

    \section{RPM Deep Learning Models}\label{sec:models}

\begin{figure}[t]
    \centering
    \includegraphics[width=.25\textwidth]{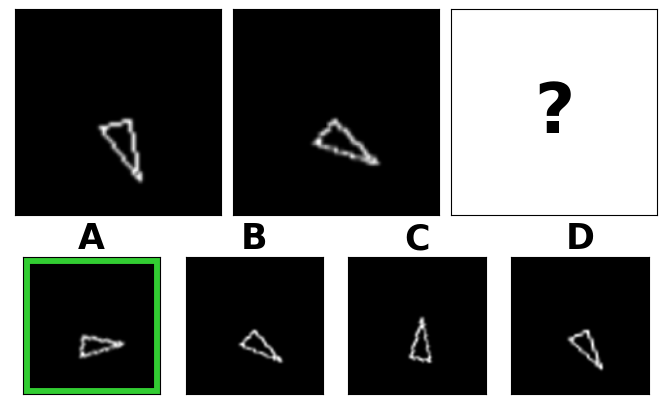}
    \caption{\textbf{RPM-like problem from~\cite{hoshen2017iq}.} The task verifies the ability to identify progression rule applied to triangle rotation. Correct answer is highlighted with a green bounding box.}
    \label{fig:rpm-hoshen2017iq}
\end{figure}

One of the first DL models for solving RPMs was proposed by
\citet{hoshen2017iq} who constructed a convolutional neural network (CNN)~\cite{lecun1990handwritten} for solving geometric pattern recognition problems, capable of generating images according to a specified pattern.
Moreover, an additional network based on the same visual backbone was employed to solve problems involving rotation, reflection, color, size and shape of the patterns (see Fig.~\ref{fig:rpm-hoshen2017iq}).
CNNs were further utilised in~\cite{mekik2018similarity}, where a generalised similarity-based approach to solving RPMs from the Sandia suite~\cite{matzen2010recreating} (cf. Fig.~\ref{fig:rpm-sandia}) was presented.
The method eliminates the need for structure mapping and instead relies on feature-based processing using relational and non-relational features.

\begin{figure}[t]
    \centering
    \includegraphics[width=.3\textwidth]{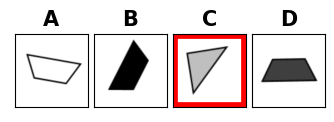}
    \caption{
    \textbf{Odd-one-out matrix.}
    Example of an odd-one-out problem from~\cite{mandziuk2019deepiq}.
    The task is to point the mismatched element out of several choices.
    In the example it is a triangle (all others are trapezoids) - marked with a red boundary.
    }
    \label{fig:odd-one-out-deepiq}
\end{figure}

Another feature-centric approach is discussed in~\cite{mandziuk2019deepiq}, where an autoencoder with dense fully-connected layers is first trained to learn feature-based representations and then combined with an ensemble of shallow multilayer perceptrons (MLPs).
On top, a scoring module is proposed that can be adjusted to the final downstream task.
\citet{mandziuk2019deepiq} have shown that this approach facilitates transfer learning, by applying the model trained to solve RPMs (cf.\ Fig.~\ref{fig:rpm-deepiq}) to other problems with a similar input distribution, e.g.\ the odd-one-out tasks (see~Fig.~\ref{fig:odd-one-out-deepiq}).
Although the above works have demonstrated that classic neural network architectures are capable of possessing abstract visual reasoning skills, the approaches were evaluated on visually-simple RPM benchmarks with limited possibility to verify out-of-distribution generalisation.

In the following sections we focus on consecutive works that considered RPMs with compositional structure and tackled o.o.d.\ generalisation challenges.
A high-level overview of DL models proposed in these works is presented in Table~\ref{tab:rpm-models}.
Besides baseline methods, we categorise the followup models into two classes: relational reasoning networks and hierarchical networks.
Models of the first type are inherently based on the Relation Network~\cite{santoro2017simple}, whereas those of the second type, inspired by the hierarchical nature of RPMs, inject various structural inductive biases into network architectures used to solve RPMs.

\begin{table*}[!ht]
    \centering
    \caption{
    \textbf{Model overviews.}
    A summary of key design choices adopted in RPM solving models.
    \emph{Dataset} column shows the works in which a given (model, dataset) pair was included in the experiments.
    In the \emph{Training setup} column, \emph{Target cross-entropy loss} refers to the supervised training approaches, whereas \emph{Meta-target binary cross-entropy loss} refers to the auxiliary training settings.}
    \begin{tabular}{p{.12\textwidth}p{.12\textwidth}p{.1\textwidth}p{.3\textwidth}p{.25\textwidth}}
        \toprule
        \rowcolor{lightergray} \textbf{Model}    & \textbf{Dataset}                                              & \textbf{Input}                             & \textbf{Design highlight}                                                                                                                                                                                  & \textbf{Training setup}                                 \\
        \midrule
        \makecell[cp{.12\textwidth}]{Baselines~\cite{barrett2018measuring,zhang2019raven}} & \makecell[cp{.12\textwidth}]{PGM~\cite{barrett2018measuring} \\ RAVEN~\cite{zhang2019raven} \\ I-RAVEN~\cite{hu2021stratified}} & Single panels & \makecell[cp{.3\textwidth}]{Embeds single panels with a shallow CNN or ResNet-18 and aggregates information from RPM context panels with an MLP or LSTM.} & \makecell[cp{.25\textwidth}]{Target cross-entropy loss \\ Meta-target binary cross-entropy loss} \\
        \midrule
        \makecell[cp{.12\textwidth}]{Context-blind models~\cite{barrett2018measuring}} & \makecell[cp{.12\textwidth}]{PGM~\cite{barrett2018measuring} \\ RAVEN~\cite{wu2020scattering,hu2021stratified} \\ I-RAVEN~\cite{wu2020scattering,hu2021stratified}} & Single panels & \makecell[cp{.3\textwidth}]{Reason only about answer panels and do not rely on RPM context.} & \makecell[cp{.25\textwidth}]{Target cross-entropy loss} \\
        \specialrule{1pt}{3pt}{3pt}
        Wild ResNet~\cite{barrett2018measuring} & \makecell[cp{.12\textwidth}]{PGM~\cite{barrett2018measuring} \\ I-RAVEN~\cite{hu2021stratified}} & \makecell[cp{.1\textwidth}]{Context} & \makecell[cp{.3\textwidth}]{Uses ResNet on a stack of 9 context panels.} & \makecell[cp{.25\textwidth}]{Target cross-entropy loss} \\
        \midrule
        WReN~\cite{barrett2018measuring} & \makecell[cp{.12\textwidth}]{PGM~\cite{barrett2018measuring} \\ RAVEN~\cite{zhang2019raven} \\ I-RAVEN~\cite{hu2021stratified}} & Single panels & \makecell[cp{.3\textwidth}]{Aggregates information from pairs of RPM context panels with the RN.} & \makecell[cp{.25\textwidth}]{Target cross-entropy loss \\ Meta-target binary cross-entropy loss} \\
        \midrule
        VAE-WReN~\cite{steenbrugge2018improving} & \makecell[cp{.12\textwidth}]{PGM~\cite{barrett2018measuring}} & Single panels & \makecell[cp{.3\textwidth}]{Replaces the CNN panel encoder of WReN with a disentangled $\beta$-VAE trained separately from the WReN model.} & \makecell[cp{.25\textwidth}]{Target cross-entropy loss \\ Meta-target binary cross-entropy loss \\ Modified ELBO objective~\cite{burgess2018understanding}
        } \\
        \midrule
        ARNe~\cite{hahne2019attention}           & \makecell[cp{.12\textwidth}]{PGM~\cite{hahne2019attention}   \\ RAVEN~\cite{hahne2019attention}}           & Single panels   & \makecell[cp{.3\textwidth}]{Extends the WReN model with the Transformer located between panel embedding component and the RN module.} & \makecell[cp{.25\textwidth}]{Target cross-entropy loss \\ Meta-target binary cross-entropy loss}           \\
        \midrule
        CoPINet~\cite{zhang2019learning}           & \makecell[cp{.12\textwidth}]{PGM~\cite{zhang2019learning}   \\ RAVEN~\cite{zhang2019learning}           \\ I-RAVEN~\cite{hu2021stratified}}           & \makecell[cp{.1\textwidth}]{Single panels}   & \makecell[cp{.3\textwidth}]{Introduces an explicit permutation-invariant contrasting neural module for distinguishing features between answer panels.} & \makecell[cp{.25\textwidth}]{Variant of NCE loss~\cite{zhang2019learning}} \\
        \midrule
        LEN~\cite{zheng2019abstract}           & \makecell[cp{.12\textwidth}]{PGM~\cite{zheng2019abstract}   \\ RAVEN~\cite{zheng2019abstract}           \\ I-RAVEN~\cite{hu2021stratified}}           & \makecell[cp{.1\textwidth}]{Single panels,   \\ Context$\setminus a_k$}           & \makecell[cp{.3\textwidth}]{Aggregates information from triples of RPM context panels and a context embedding vector with the RN.}   & \makecell[cp{.25\textwidth}]{Target cross-entropy loss \\ Meta-target binary cross-entropy loss           \\ Feature Robust Abstract Reasoning}           \\
        \midrule
        SRAN~\cite{hu2021stratified}           & \makecell[cp{.12\textwidth}]{PGM~\cite{hu2021stratified}   \\ I-RAVEN~\cite{hu2021stratified}}           & \makecell[cp{.1\textwidth}]{Single panels,   \\ Rows, Cols,           \\ Pairs of rows           \\ Pairs of cols}           & \makecell[cp{.3\textwidth}]{Gradually aggregates features from panel hierarchies with a gated embedding fusion module.}   & \makecell[cp{.25\textwidth}]{($N$+$1$)-tuplet loss~\cite{sohn2016improved}} \\
        \midrule
        MXGNet~\cite{wang2020abstract}           & \makecell[cp{.12\textwidth}]{PGM~\cite{wang2020abstract}   \\ RAVEN~\cite{wang2020abstract}}           & \makecell[cp{.1\textwidth}]{Single panels}   & \makecell[cp{.3\textwidth}]{Reasons about inter-panel relations with multiplex graph neural networks.} & \makecell[cp{.25\textwidth}]{Target cross-entropy loss \\ Meta-target binary cross-entropy loss}           \\
        \midrule
        SCL~\cite{wu2020scattering}           & \makecell[cp{.12\textwidth}]{PGM~\cite{wu2020scattering,malkinski2020multilabel}   \\ RAVEN~\cite{wu2020scattering}           \\ I-RAVEN~\cite{wu2020scattering,malkinski2020multilabel}}           & \makecell[cp{.1\textwidth}]{Single panels}   & \makecell[cp{.3\textwidth}]{Reasons about inter-panel and intra-panel relations with a neural module based on a scattering transformation.} & \makecell[cp{.25\textwidth}]{Target cross-entropy loss} \\
        \midrule
        DCNet~\cite{zhuo2021effective}           & \makecell[cp{.12\textwidth}]{PGM~\cite{zhuo2021effective}   \\ RAVEN~\cite{zhuo2021effective}}           & \makecell[cp{.1\textwidth}]{Rows   \\ Cols}           & \makecell[cp{.3\textwidth}]{Considers similarity of third row/col to the first and second ones and differences in the last row completed by each answer panel. }   & \makecell[cp{.25\textwidth}]{Target binary cross-entropy loss} \\
        \midrule
        MRNet~\cite{benny2020scale}           & \makecell[cp{.12\textwidth}]{PGM~\cite{benny2020scale}   \\ RAVEN~\cite{benny2020scale}           \\ RAVEN-FAIR~\cite{benny2020scale}}           & \makecell[cp{.1\textwidth}]{Single panels}   & \makecell[cp{.3\textwidth}]{Applies separate RNs to panel embeddings in three different resolutions.} & \makecell[cp{.25\textwidth}]{Weighted target binary cross-entropy loss} \\
        \midrule
        Rel-Base~\cite{spratley2020closer}           & \makecell[cp{.12\textwidth}]{PGM~\cite{spratley2020closer}   \\ RAVEN~\cite{spratley2020closer}}           & \makecell[cp{.1\textwidth}]{Single panels}   & \makecell[cp{.3\textwidth}]{Processes a stack of 9 context panel embeddings with a 1D convolution module.} & \makecell[cp{.25\textwidth}]{Target cross-entropy loss} \\
        \midrule
        Rel-AIR~\cite{spratley2020closer}           & \makecell[cp{.12\textwidth}]{PGM~\cite{spratley2020closer}   \\ RAVEN~\cite{spratley2020closer}}           & \makecell[cp{.1\textwidth}]{Single panels}   & \makecell[cp{.3\textwidth}]{Reasons about segmented objects obtained with AIR unsupervised scene decomposition model.} & \makecell[cp{.25\textwidth}]{Target cross-entropy loss \\ Unsupervised scene decomposition pre-training} \\
        \midrule
        MLRN~\cite{jahrens2020solving}           & \makecell[cp{.12\textwidth}]{PGM~\cite{jahrens2020solving}}   & \makecell[cp{.1\textwidth}]{Single panels} & \makecell[cp{.3\textwidth}]{Extends WReN with a Multi-Layer Relation Network, encodes the input images with a Magnitude Encoding and uses the LAMB optimiser.} & \makecell[cp{.25\textwidth}]{Target cross-entropy loss} \\
        \midrule
        PrAE~\cite{zhang2021abstract}           & \makecell[cp{.12\textwidth}]{RAVEN~\cite{zhang2021abstract}}  & \makecell[cp{.1\textwidth}]{Single panels} & \makecell[cp{.3\textwidth}]{Neuro-symbolic approach that produces a probabilistic RPM representation and uses it to generate probable answer via probabilistic abduction and execution.} & \makecell[cp{.25\textwidth}]{REINFORCE~\cite{williams1992simple}} \\
        \midrule
        NI~\cite{rahaman2021dynamic} & \makecell[cp{.12\textwidth}]{PGM~\cite{rahaman2021dynamic}} & \makecell[cp{.1\textwidth}]{Single panels} & \makecell[cp{.3\textwidth}]{Compositional model comprising reusable self-attention layers with a learnable routing mechanism.} & \makecell[cp{.25\textwidth}]{Target cross-entropy loss} \\
        \bottomrule
    \end{tabular}
    \label{tab:rpm-models}
\end{table*}

\subsection{Baselines}
Similarly to~\cite{hoshen2017iq}, baseline models for solving RPMs are formed by using a CNN as a visual feature extractor, followed by a reasoning module.
The perceptual backbone often consists of 4 convolution layers with non-linearities in-between or a variant of ResNet~\cite{he2016deep} with 18, 34 or 50 layers.
In simple baseline models~\cite{barrett2018measuring,zhang2019raven}, the visual component processes each image independently.
Then, the features extracted by these visual modules from all matrix panels are either concatenated into a single vector and fed into an MLP or stacked to form a time-series and processed by an LSTM~\cite{hochreiter1997long}.
Nonetheless, it was shown that such typical neural architectures struggle even in simpler RPM generalisation regimes~\cite{barrett2018measuring,zhang2019raven}, which raises the need for dedicated AVR models.

\begin{figure*}[t]
    \centering
    \subfloat[Single panel.]{
        \includegraphics[width=.12\textwidth]{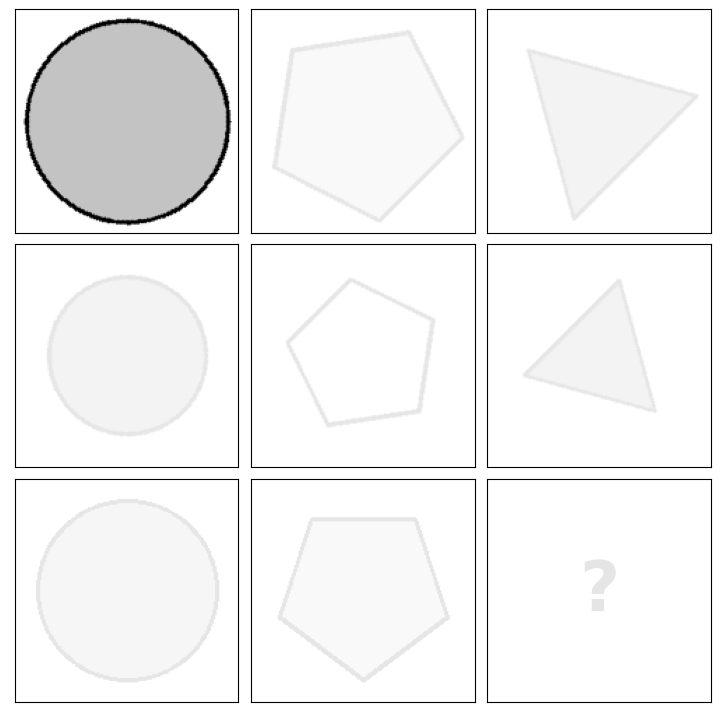}
        \label{fig:rpm-panel-hierarchies-single-panel}
    }
    \hfil
    \subfloat[Single row.]{
        \includegraphics[width=.12\textwidth]{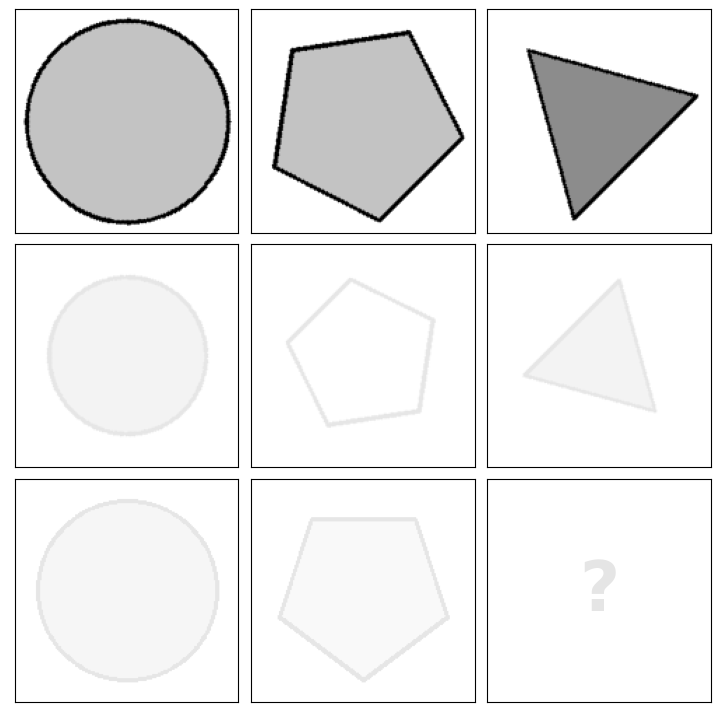}
        \label{fig:rpm-panel-hierarchies-single-row}
    }
    \hfil
    \subfloat[Single col.]{
        \includegraphics[width=.12\textwidth]{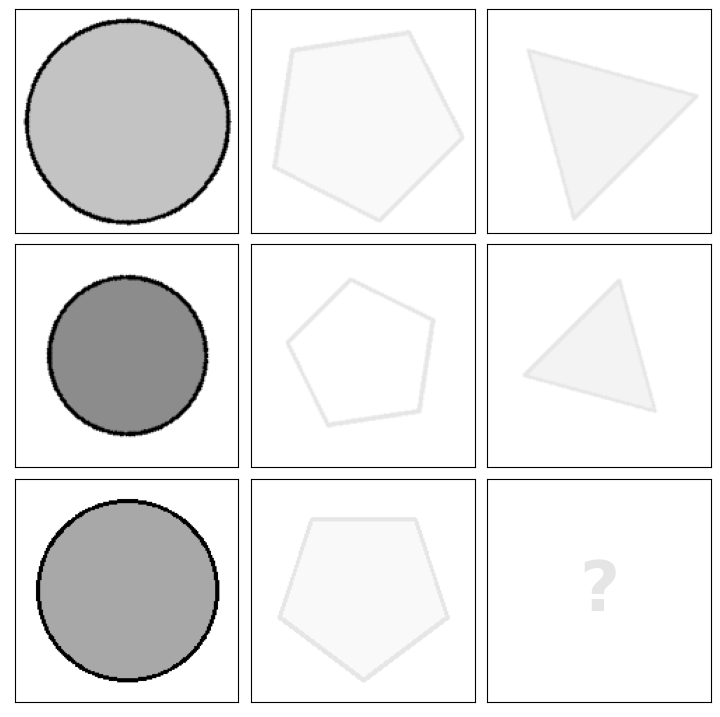}
        \label{fig:rpm-panel-hierarchies-single-col}
    }
    \hfil
    \subfloat[Pair of rows.]{
        \includegraphics[width=.12\textwidth]{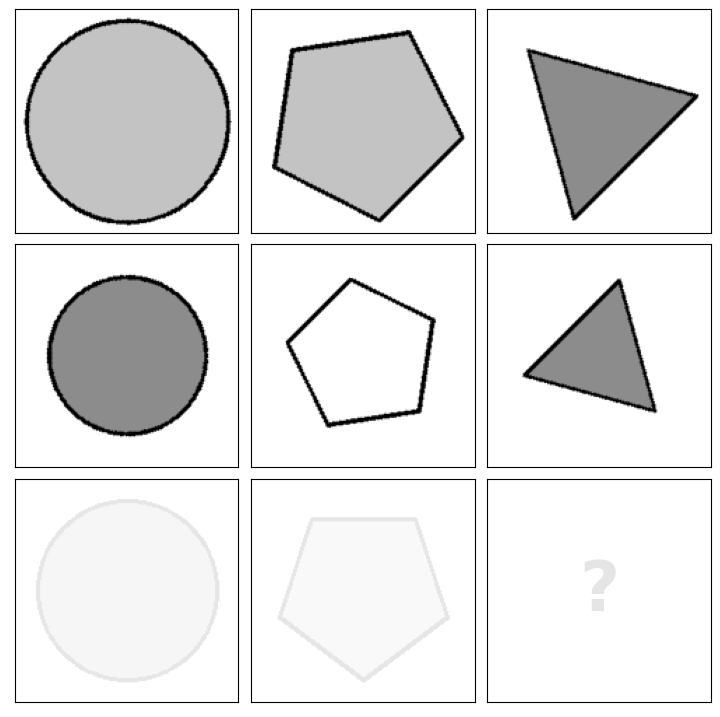}
        \label{fig:rpm-panel-hierarchies-pair-of-rows}
    }
    \hfil
    \subfloat[Pair of Cols.]{
        \includegraphics[width=.12\textwidth]{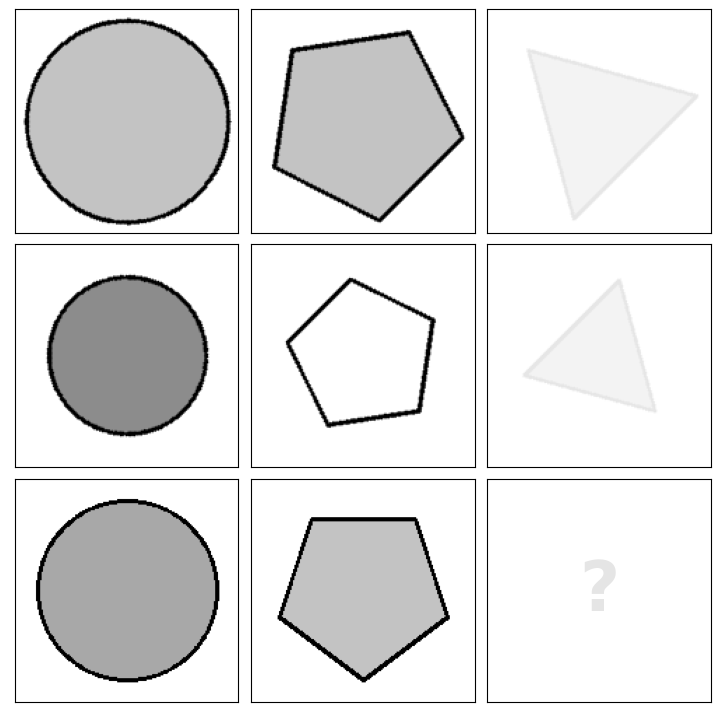}
        \label{fig:rpm-panel-hierarchies-pair-of-cols}
    }
    \hfil
    \subfloat[Context$\setminus a_k$.]{
        \includegraphics[width=.12\textwidth]{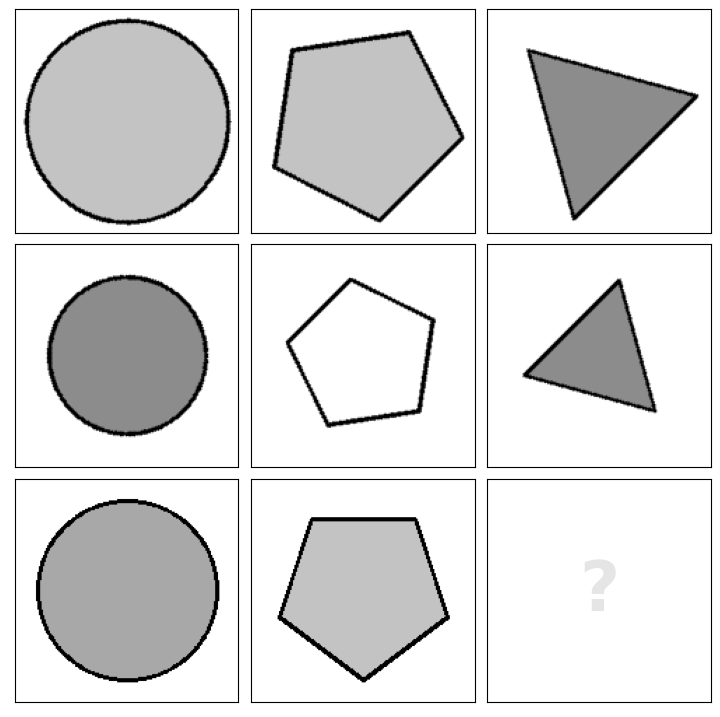}
        \label{fig:rpm-panel-hierarchies-context}
    }
    \hfil
    \subfloat[Context.]{
        \includegraphics[width=.12\textwidth]{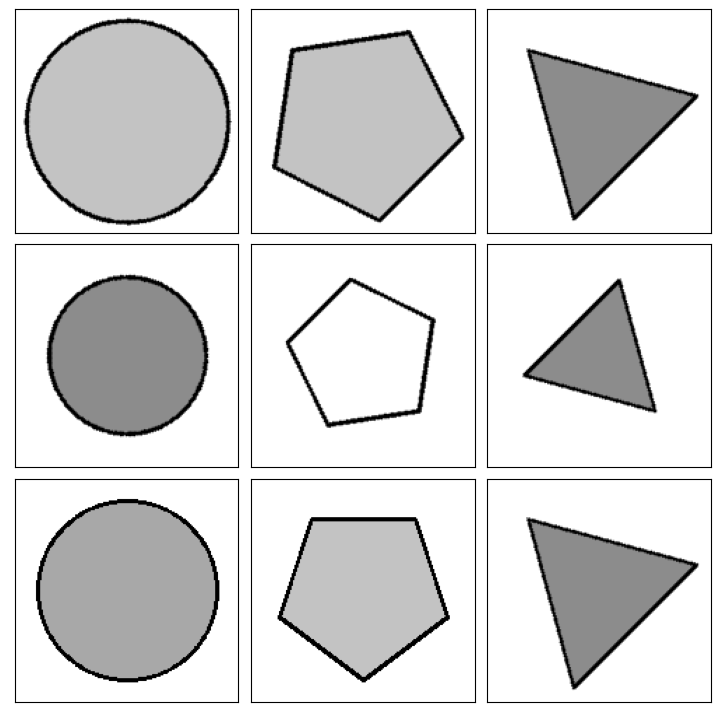}
        \label{fig:rpm-panel-hierarchies-completed-context}
    }
    \caption{
    \textbf{RPM panel hierarchies.}
    Each hierarchy demonstrates how panels can be jointly processed.
    For instance, when building hierarchical perceptual model backbones (e.g. in SRAN~\cite{hu2021stratified}), panels are stacked on top of each other to form a matrix of the shape $(c \times w \times h)$, where $c$ is the number of panels in the given hierarchy, $w$ - the image width and $h$ - the image height.
    Alternatively, the embeddings of these panels can be combined in analogous manner (e.g. in LEN~\cite{zheng2019abstract}) and concatenated into a single vector -- a representation of given hierarchy.
    The hierachy which contains 8 RPM context images without the remaining missing panel is denoted as ``Context$\setminus a_k$'', whereas the hierarchy with full RPM context where one of the answers is placed in the bottom-right panel as ``Context''.
    Additional hierarchies may include diagonal panels (e.g.~\cite{mandziuk2019deepiq,wang2020abstract}) or random combinations (e.g.~\cite{zheng2019abstract,wang2020abstract}).
    }
    \label{fig:rpm-panel-hierarchies}
\end{figure*}

\subsection{Relational reasoning networks}
Seminal works that introduced challenging RPM reasoning benchmarks (PGM~\cite{barrett2018measuring} and RAVEN~\cite{zhang2019raven}) have shown that baseline DL models generally lack the relational reasoning capability that is inherent to the AVR domain.
Similar observations were noted in other (non AVR) problems involving relational reasoning that tried to tackle both artificial~\cite{johnson2017clevr,santoro2017simple} and real-world~\cite{malinowski2014multi,weston2015towards,antol2015vqa,gao2015you,ren2015image,ren2015exploring,krishna2017visual,hudson2019gqa,teney2020v} challenges.
With the aim of equipping neural modules with relational reasoning abilities, \citet{santoro2017simple} proposed the Relation Network (RN) -- a simple neural module for relational reasoning.
RN arranges a set of objects $O$ into pairs (where an object $o \in O$ can be generally represented by any vector) and summarises the whole set of objects into a single descriptive representation with
\[
    RN(O) = f_{\phi}(\sum_{i,j}g_\theta(o_i, o_j))
\]
where $f_\phi$ and $g_\theta$ are functions typically implemented as MLPs.
\citet{santoro2017simple} successfully applied RN to problems from diverse domains including VQA~\cite{johnson2017clevr}, text-based question answering~\cite{weston2015towards} and reasoning about dynamic physical systems~\cite{santoro2017simple}.
The module was later extended in several ways, e.g.\ to recurrent version capable of sequential relational reasoning~\cite{palm2018recurrent}, and its learning capacity was enhanced by stacking multiple layers~\cite{jahrens2019multi}.
    As a follow-up to~\cite{santoro2017simple}, in~\cite{barrett2018measuring} the RN was integrated as part of an end-to-end architecture for solving RPMs -- the Wild Relation Network (WReN).
The model outperformed baselines by a significant margin and sparked interest within the AVR community in building models that employ RN\@.

Multiple attempts have been made to either extend the originally proposed WReN model or to incorporate the RN into another end-to-end architecture.
\citet{steenbrugge2018improving} introduced the VAE-WReN, which replaced the CNN backbone of WReN with a disentangled variational autoencoder~\cite{kingma2013auto,higgins2016beta,burgess2018understanding} and showed improved generalisation when tested on PGM\@.
The RN part of WReN was further extended to a multi-layer version~\cite{jahrens2019multi}.
The Multi-Layer Relation Network (MLRN) combined with $\ell_2$-regularization, Magnitude Encoding and the LAMB optimiser~\cite{you2020large} presented close to perfect performance in neutral PGM regime, however, its performance was worse than that of VAE-WReN in other generalisation regimes, suggesting that the model was prone to overfitting.

Inspired by advances in psychology, which suggest that attention mechanisms play a crucial role in human visual reasoning capabilities, the Attention Relation Network (ARNe) was proposed in~\cite{hahne2019attention}.
ARNe builds on WReN and equips it with attention mechanism borrowed from the Transformer~\cite{vaswani2017attention}.
However, the increased model complexity was not fully justified, as the model demonstrated small improvements on PGM and performed worse than baselines on RAVEN\@.
Nonetheless, \citet{hahne2019attention} have shown that after increasing RAVEN dataset size 5-fold, ARNe performance increased substantially, which suggests that the proposed attention mechanism for solving AVR tasks is promising, despite being inefficient w.r.t.\ the sample size.
Possible directions for future work in this area may involve further research on incorporating the attention mechanism, which is additionally motivated by the impressive performance of attention-based models in other domains, e.g.~\cite{vaswani2017attention,dong2018speech,dosovitskiy2021an,jaegle2021perceiver}.

In the original WReN proposal~\cite{barrett2018measuring}, the relational module operated on pairs of panel embeddings.
However, as described in~\cite{carpenter1990one}, the rules in RPMs are inherently applied row- or column-wise.
Therefore, some attempts have been made to incorporate this structural bias in the RN module.
\citet{benny2020scale} proposed the MRNet, a model which first generates panel embeddings in three different resolutions and then processes them with distinct RN modules that consider triples of embeddings from each RPM row and column, respectively.
In a similar spirit,~\citet{zheng2019abstract} introduced the Logic Embedding Network (LEN) -- a model with several improvements to WReN in the context of solving RPMs.
Firstly, the approach arranges RPM context panel embeddings into triples instead of pairs.
The triples include row- and column-wise combinations (similarly to~\cite{benny2020scale}, although, only a single resolution is considered) processed by a neural module $g$ with parameters $\theta_1$.
In contrast to~\cite{benny2020scale}, LEN additionally processes the remaining possible panel embedding combinations with a parallel module of the same structure as $g$, but with different parameters $\theta_2$.
Moreover, with each triple a representation of the whole RPM context is concatenated.
This representation is obtained with a supplementary CNN that processes a stack of 8 context panels.
This structural inductive bias allows to reason about the relations between multiple panels more accurately and forms a basis for a group of hierarchical models discussed in the following section.

At the intersection of relational reasoning and hierarchical networks, another model---the Contrastive Perceptual Inference Network (CoPINet)~\cite{zhang2019learning}---is located.
It is a permutation-invariant approach with an explicit contrastive mechanism that helps to distinguish the correct RPM answer.
CoPINet first extracts independent panel embeddings with a visual backbone and then iteratively applies the contrastive module capable of discovering features that differentiate among a set of choices.
In contrast to RN that considers pairs of feature embeddings, CoPINet's contrasting mechanism collates each object representation with an aggregated representation of all remaining objects.
The initial object representation is obtained by summing outputs of the visual backbone for each image along RPM's rows and columns, respectively.
Next, these representations are iteratively refined using the contrast module.
Although CoPINet was shown to possess impressive reasoning ability on matrices from RAVEN~\cite{zhang2019learning}, it was later discovered that the model performs sub-par on the balanced version of this dataset (I-RAVEN)~\cite{hu2021stratified}.
Therefore, additional validation of CoPINet's architecture in other AVR problems sets an interesting avenue for future work.

\subsection{Hierarchical networks}
In multiple areas where DL models thrive, it has often been beneficial to incorporate domain-specific knowledge about the problem structure into the network architecture.
Notable examples include making CNNs translation-invariant in problems with 2D images~\cite{kayhan2020translation} or rotation-equivariant for spherical images~\cite{cohen2018spherical}.
As demonstrated by the LEN model~\cite{zheng2019abstract}, such inductive biases are also helpful in solving RPMs, e.g.\ by specifying in what manner panel representations should be processed.
Figure~\ref{fig:rpm-panel-hierarchies} summarises the most common hierarchies utilised by RPM DL models.
While LEN exploited a single row (Fig.~\ref{fig:rpm-panel-hierarchies-single-row}) and a single column (Fig.~\ref{fig:rpm-panel-hierarchies-single-col}) hierarchies, respectively together with the 8 context panels (Fig.~\ref{fig:rpm-panel-hierarchies-context}), subsequent DL approaches relied on additional techniques.

While solving RPMs, it is often required not only to identify a set of rules that govern a single row/column but also to subsequently find an analogous set of rules applied to another row/column from the same matrix.
Motivated by this observation, the Stratified Rule-Aware Network (SRAN)~\cite{hu2021stratified} devotes particular attention to pairs of rows (Fig.~\ref{fig:rpm-panel-hierarchies-pair-of-rows}) and pairs of columns (Fig.~\ref{fig:rpm-panel-hierarchies-pair-of-cols}).
From each such pair that includes 6 images, the panels are stacked on top of each other and processed by a dedicated convolutional pathway.
SRAN gradually aggregates representations from consecutive hierarchies (single panel, single row/col, pair of rows/cols) using a gated embedding fusion module realised by an MLP\@.

Such a gradual processing of latent features was also shown to be beneficial in the Scattering Compositional Learner (SCL) model~\cite{wu2020scattering}.
Similarly to SRAN, SCL first computes embeddings for each RPM panel.
Then, the matrix is iteratively completed by one of the answer panels.
In each step, the model computes a joint representation of all 9 context images.
The obtained embedding is finally fed to a scoring module which produces a probablity of correctness of the selected answer.
In contrast to SRAN, SCL replaces MLP components with a scattering transformation that splits the input into multiple groups, applies the same neural module to each group, and merges the outputs into a single embedding.
The approach shares some ideas with group convolution~\cite{krizhevsky2012imagenet}, ResNeXt~\cite{xie2017aggregated} and Modular Networks~\cite{kirsch2018modular}.

Processing order similar to SCL was employed in Rel-Base and Rel-AIR models~\cite{spratley2020closer} that also start with building embeddings independently for each matrix panel.
The models differ from SCL with the choice of the encoder network -- instead of a combination of CNN and the scattering transformation they use shallow ResNet.
Next, both models aggregate 9 context embeddings.
However, in contrast to SCL, these representations are processed in Rel-Base/AIR models with a simple 1D convolution rather than the scattering transformation.
In this view, Rel-Base and Rel-AIR relate to WReN which instead of 1D convolution uses more computationally expensive RN\@.
Although the models proposed by~\citet{spratley2020closer} are structurally similar, they operate on different inputs.
Rel-Base processes original matrix panels, whereas Rel-AIR employs an unsupervised scene decomposition pre-processing step with the Attend-Infer-Repeat (AIR)~\cite{eslami2016attend} neural model.
AIR decomposes each panel into several object slots that are used to compute a panel embedding.
This allows to disentangle single objects from the overall scene and simplifies the process of discovering relations between them.
Explicit scene representation is additionally proven to be useful in~\cite{zhang2021abstract}, where it is represented in a probabilistic manner.
The authors combine this abstract scene representation with a scene inference engine to produce plausible answers to RPMs with a neuro-symbolic approach.

In~\cite{zhuo2021effective}, similarly to SRAN, the authors further build on the importance of instantiating RPM rules row- and column-wise and propose a Dual-Contrast Network (DCNet).
The model uses a rule contrast module that compares representations of the completed third row with the first and second row representations (resp.\ for columns) already present in the RPM\@.
DCNet additionally incorporates a contrast module similar to that of CoPINet, which increases relative differences between the candidate embeddings.
The model training algorithm resembles that of SRAN, where the similarity of the representation of the first two rows/columns from the RPM context to embeddings of row/column pairs including the correct answer panel is maximised, while the similarity to embeddings of row/column pairs with wrong answer panels is minimised.

RPM hierarchies were further exploited in~\cite{wang2020abstract}, where the authors propose MXGNet -- a deep graph neural network for solving AVR problems.
In contrast to the already described approaches where structural inductive biases were hand-crafted, MXGNet builds on an adaptive mechanism which automatically selects key problem hierarchies depending on the task at hand.
This is achieved by employing $\ell_1$-regularised gating variables that measure to which extent the panel subsets contribute to the model performance.
The authors found out that indeed single row/column hierarchies are crucial for high model performance in solving RPMs.

Another unique approach was presented in~\cite{rahaman2021dynamic}, where the authors propose a method of dynamic inference with Neural Interpreters (NI).
The model is composed of several reusable self-attention blocks with a learnable routing mechanism.
The method is inspired by the design of programming languages and utilizes concepts analogous to scripts, functions, variables, and an interpreter.
This perspective facilitates compositional reasoning, where model blocks can be reused across tasks.
Another design highlight of the model lies in the input processing module -- the model splits each RPM panel into smaller patches instead of taking the whole images as inputs.
This division reduces input dimensionality, which allows to apply even computationally expensive modules -- the authors employ the attention module, which computes pair-wise interactions between input elements (image pixels).
The approach of splitting an image into patches was initially proposed in the Vision Transformer (ViT)~\cite{dosovitskiy2021an}, which~\citet{rahaman2021dynamic} adapted to solve RPMs from PGM.

    \begin{table*}[t]
    \centering
    \caption{
    \textbf{PGM accuracy.}
    Accuracy in all regimes of the PGM dataset~\cite{barrett2018measuring} arranged in the ascending order by score on the test set of the Neutral regime.
    The Held-out Attribute Pairs regime is denoted as H.O.\ A.P., Held-out Triple Pairs as H.O.\ T.P., Held-out Triples as H.O.\ Triples, Held-out Attribute \texttt{line-type} as H.O.\ L-T and Held-out Attribute \texttt{shape-colour} as H.O.\ S-C.
    }
    \scriptsize
    \begin{tabular}{l|cc|cc|cc|cc|cc|cc|cc|cc}
        \toprule
        \multirow{3}{*}{Method} & \multicolumn{16}{c}{Accuracy (\%)} \\
        & \multicolumn{2}{c|}{Neutral} & \multicolumn{2}{c|}{Interpolation} & \multicolumn{2}{c|}{H.O.\ A.P.} & \multicolumn{2}{c|}{H.O.\ T.P.} & \multicolumn{2}{c|}{H.O. Triples} & \multicolumn{2}{c|}{H.O.\ L-T} & \multicolumn{2}{c|}{H.O.\ S-C} & \multicolumn{2}{c}{Extrapolation} \\
        & Val. & Test.         & Val. & Test.         & Val. & Test.         & Val. & Test.         & Val. & Test.         & Val. & Test.         & Val. & Test.         & Val. & Test.         \\
        \midrule
        Context-blind ResNet~\cite{barrett2018measuring}   & -    & 22.4          & -    & -             & -    & -             & -    & -             & -    & -             & -    & -             & -    & -             & -    & -             \\
        CNN MLP~\cite{barrett2018measuring}                & -    & 33.0          & -    & -             & -    & -             & -    & -             & -    & -             & -    & -             & -    & -             & -    & -             \\
        CNN LSTM~\cite{barrett2018measuring}               & -    & 35.8          & -    & -             & -    & -             & -    & -             & -    & -             & -    & -             & -    & -             & -    & -             \\
        ResNet-50~\cite{barrett2018measuring}              & -    & 42.0          & -    & -             & -    & -             & -    & -             & -    & -             & -    & -             & -    & -             & -    & -             \\
        NCD~\cite{zhuo2021unsupervised}            & -    & 47.6          & -    & 47.0             & -    & -             & -    & -             & -    & -             & -    & -             & -    & -             & -    & \textbf{24.9}             \\
        Wild-ResNet~\cite{barrett2018measuring}            & -    & 48.0          & -    & -             & -    & -             & -    & -             & -    & -             & -    & -             & -    & -             & -    & -             \\
        CoPINet~\cite{zhang2019learning}                   & -    & 56.4          & -    & -             & -    & -             & -    & -             & -    & -             & -    & -             & -    & -             & -    & -             \\
        WReN $\beta=0$~\cite{barrett2018measuring}         & 63.0 & 62.6          & 79.0 & 64.4          & 46.7 & 27.2          & 63.9 & 41.9          & 63.4 & 19.0          & 59.5 & 14.4          & 59.1 & 12.5          & 69.3 & 17.2          \\
        VAE-WReN $\beta=4$~\cite{steenbrugge2018improving} & 64.8 & 64.2          & -    & -             & 70.1 & 36.8          & 64.6 & 43.6          & 59.5 & 24.6          & -    & -             & -    & -             & -    & -             \\
        MXGNet $\beta=0$~\cite{wang2020abstract}           & 67.1 & 66.7          & 74.2 & 65.4          & 68.3 & 33.6          & 67.1 & 43.3          & 63.7 & 19.9          & 60.1 & 16.7          & 68.5 & 16.6          & 69.1 & 18.9          \\
        LEN $\beta=0$~\cite{zheng2019abstract}             & -    & 68.1          & -    & -             & -    & -             & -    & -             & -    & -             & -    & -             & -    & -             & -    & -             \\
        DCNet~\cite{zhuo2021effective}                     & -    & 68.6          & -    & 59.7          & -    & -             & -    & -             & -    & -             & -    & -             & -    & -             & -    & 17.8          \\
        T-LEN $\beta=0$~\cite{zheng2019abstract}           & -    & 70.3          & -    & -             & -    & -             & -    & -             & -    & -             & -    & -             & -    & -             & -    & -             \\
        SCL MLCL~\cite{malkinski2020multilabel}            & 71.0 & 71.1          & 93.2 & 70.9          & 79.7 & 66.0          & 86.1 & 71.7          & 84.0 & 22.1          & 86.1 & 16.1          & 94.1 & 12.8          & 83.0 & 21.9 \\
        SRAN~\cite{hu2021stratified}                       & -    & 71.3          & -    & -             & -    & -             & -    & -             & -    & -             & -    & -             & -    & -             & -    & -             \\
        ViT~\cite{rahaman2021dynamic} & 73.3 & 72.7          & 89.9 & 67.7 & 69.4 & 34.1 & 67.6 & 44.1 & 73.8 & 15.9 & -    & -             & -    & -             & 92.2 & 16.4 \\
        WReN $\beta=10$~\cite{barrett2018measuring}        & 77.2 & 76.9          & 92.3 & 67.4          & 73.4 & 51.7          & 74.5 & 56.3          & 80.0 & 20.1          & 78.1 & 16.4          & 85.2 & 13.0          & 93.6 & 15.5          \\
        NI~\cite{rahaman2021dynamic} & 77.3 & 77.0 & 87.9 & 70.5 & 69.5 & 36.6 & 68.6 & 45.2 & 79.9 & 20.0 & -    & -             & -    & -             & 91.8 & 19.4 \\
        WReN $\beta=10$ + TM~\cite{zheng2019abstract}      & -    & 77.8          & -    & -             & -    & -             & -    & -             & -    & -             & -    & -             & -    & -             & -    & -             \\
        LEN $\beta=10$~\cite{zheng2019abstract}            & -    & 82.3          & -    & -             & -    & -             & -    & -             & -    & -             & -    & -             & -    & -             & -    & -             \\
        T-LEN $\beta=10$~\cite{zheng2019abstract}          & -    & 84.1          & -    & -             & -    & -             & -    & -             & -    & -             & -    & -             & -    & -             & -    & -             \\
        Rel-Base~\cite{spratley2020closer}                 & -    & 85.5          & -    & -             & -    & -             & -    & -             & -    & -             & -    & -             & -    & -             & -    & 22.1             \\
        SCL CE~\cite{malkinski2020multilabel}              & 86.2 & 85.6          & 91.2 & 55.8          & 56.4 & 40.8          & 78.2 & 64.5          & 78.6 & 27.0          & 87.6 & 15.1          & 96.9 & 12.7          & 96.3 & 17.3          \\
        LEN $\beta=10$ + TM~\cite{zheng2019abstract}       & -    & 85.8          & -    & -             & -    & -             & -    & -             & -    & -             & -    & -             & -    & -             & -    & -             \\
        SCL AUX-dense~\cite{malkinski2020multilabel}       & 87.6 & 87.1          & 97.9 & 56.0          & 88.6 & \textbf{79.6} & 88.7 & \textbf{76.6} & 88.1 & 23.0          & 87.9 & 14.1          & 98.3 & 12.6          & 99.1 & 19.8          \\
        SCL AUX-sparse~\cite{malkinski2020multilabel}      & 87.4 & 87.1          & 88.1 & 54.1          & 80.8 & 63.6          & 77.5 & 64.0          & 86.0 & \textbf{30.8} & 93.5 & 17.0          & 98.0 & 12.7          & 96.9 & 17.5          \\
        ARNe $\beta=10$~\cite{hahne2019attention}          & -    & 88.2          & -    & -             & -    & -             & -    & -             & -    & -             & -    & -             & -    & -             & 98.9 & 17.8          \\
        T-LEN $\beta=10$ + TM~\cite{zheng2019abstract}     & -    & 88.9          & -    & -             & -    & -             & -    & -             & -    & -             & -    & -             & -    & -             & -    & -             \\
        SCL~\cite{wu2020scattering}                        & -    & 88.9          & -    & -             & -    & -             & -    & -             & -    & -             & -    & -             & -    & -             & -    & -             \\
        MXGNet $\beta=10$~\cite{wang2020abstract}          & 89.9 & 89.6          & 91.5 & \textbf{84.6} & 81.9 & 69.3          & 78.1 & 64.2          & 80.5 & 20.2          & 85.2 & 16.8          & 89.2 & 15.6          & 94.3 & 18.4          \\
        MRNet~\cite{benny2020scale}                        & -    & 93.4          & -    & 68.1          & -    & 38.4          & -    & 55.3          & -    & 25.9          & -    & \textbf{30.1} & -    & \textbf{16.9} & -    & 19.2          \\
        MLRN~\cite{jahrens2020solving}                     & -    & \textbf{98.0} & -    & 57.8          & -    & -             & -    & -             & -    & -             & -    & -             & -    & -             & -    & 14.9          \\
        \bottomrule
    \end{tabular}
    \label{tab:pgm}
\end{table*}

\section{Evaluating machine intelligence with RPMs}\label{sec:results}

In spite of many attempts at creating efficient models for solving RPMs, current approaches described in the previous section still struggle in more demanding benchmark setups.
In this section, we summarise the main quantitative results of the discussed models on PGM, RAVEN and I-RAVEN datasets.
Moreover, we highlight the most challenging setups for the current models and contrast their performance with the number of trainable parameters.

\subsection{Results on PGM}
We start by discussing the results on PGM\@.
Table~\ref{tab:pgm} compares performance of all discussed approaches based on the accuracy results reported in the respective papers.
Firstly, it can be seen that the majority of presented approaches weren't evaluated in all PGM regimes.
Namely, out of 32 methods shown in the table, only 9 were tested in each regime (4 unique models: WReN, MXGNet, SCL, MRNet, trained with different setups) and only 8 other methods were evaluated in at least one regime other than Neutral.

This observation raises a concern that, in practice, the PGM dataset is not utilized in a way it was intended for.
\citet{barrett2018measuring} defined the PGM benchmark as a tool for measuring generalisation performance across different regimes, whereas less than half of existing methods were evaluated on regimes other than Neutral.
In the Neutral regime, all dataset splits (train/val/test) contain RPMs with objects, attributes and rules sampled from the same underlying distributions.
As a consequence, the regime measures the capacity of the algorithm of understanding what an RPM is, rather than its ability to generalise to novel settings represented by the remaining regimes.

Although this limited evaluation of various works is partially explained by the extensive dataset size which is the main bottleneck for measuring generalisation, our beliefs are in line with the PGM authors that, actually, the performance across different regimes is what should be compared.
The accuracy in the Neutral regime is only an initial indicator of the model reasoning capability, which can often be misleading as demonstrated by variation in the MLRN performance -- the model achieves near perfect accuracy in the Neutral regime, while completely failing in the Extrapolation split.
At the same time it performs subpar when compared to other models---potentially weaker when judging by Neutral regime scores---in the Interpolation regime.
We believe that the ultimate goal of developing DL methods for solving AVR problems is not to obtain well-performing models on existing benchmarks, but rather to search for higher level approaches tackling generalisation that could be transferred to other domains.

\begin{table}[t]
    \centering
    \caption{\textbf{PGM mean accuracy.} Average accuracy measured on the test split of models that were evaluated in each PGM regime, sorted in increasing order.}
    \label{tab:pgm-mean-accuracy}
    \begin{tabular}{l|c}
        \toprule
        Method                                        & Mean test accuracy (\%) \\
        \midrule
        WReN $\beta=0$~\cite{barrett2018measuring}    & 32.4                    \\
        MXGNet $\beta=0$~\cite{wang2020abstract}      & 35.1                    \\
        WReN $\beta=10$~\cite{barrett2018measuring}   & 39.7                    \\
        SCL CE~\cite{malkinski2020multilabel}         & 39.8                    \\
        SCL AUX-sparse~\cite{malkinski2020multilabel} & 43.3                    \\
        MRNet~\cite{benny2020scale}                   & 43.4                    \\
        SCL MLCL~\cite{malkinski2020multilabel}       & 44.1                    \\
        SCL AUX-dense~\cite{malkinski2020multilabel}  & 46.1                    \\
        MXGNet $\beta=10$~\cite{wang2020abstract}     & \textbf{47.3}           \\
        \bottomrule
    \end{tabular}
\end{table}

A closer look at the models' performance in different regimes reveals that there isn't a single method that performs superior across all regimes.
On the contrary, each method seems to possess specific generalisation capabilities.
The average accuracy of the models that were evaluated in all PGM regimes is summarised in Table~\ref{tab:pgm-mean-accuracy}.
Best result is achieved by MXGNet trained using auxiliary training with $\beta=10$.
The SCL model achieves competitive results with various training setups.
The third model, MRNet is a close runner-up to SCL\@.
The WReN model performs worst.
However, since only a handful of the proposed methods were evaluated in each regime, it is difficult to ultimately choose a model with the best generalisation performance.

\begin{figure}[t]
    \centering
    \includegraphics[width=.485\textwidth]{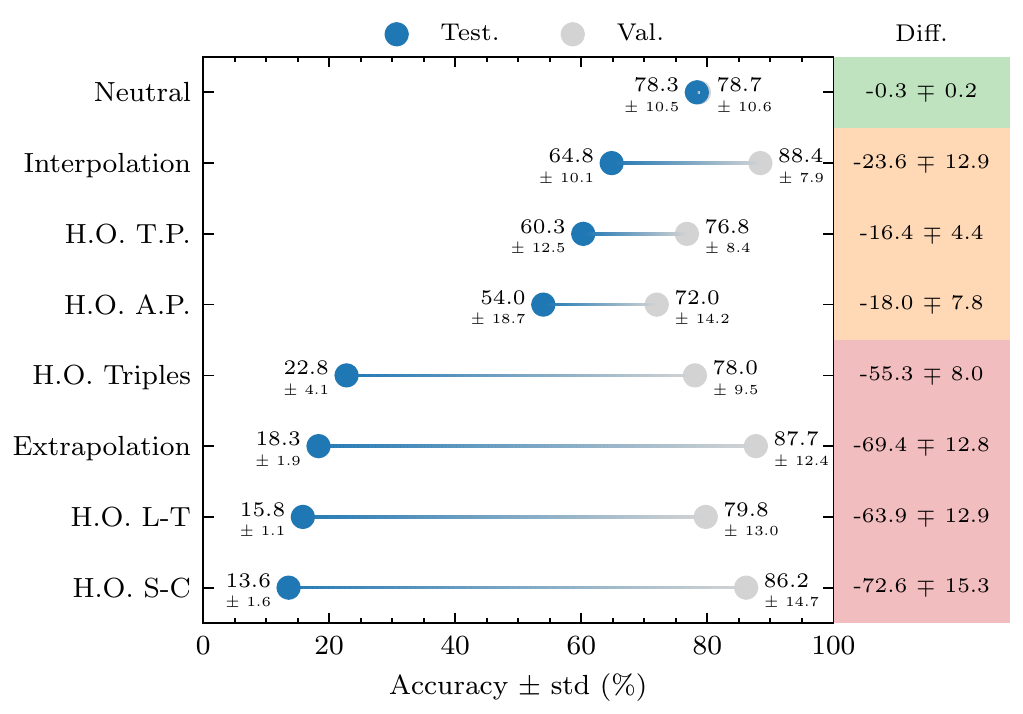}
    \caption{\textbf{PGM regime difficulty.} Mean accuracy on test (Test.) and validation (Val.) splits and their difference (Diff.) for the methods that were evaluated in each regime (listed in Table~\ref{tab:pgm-mean-accuracy}).}
    \label{fig:pgm-regime-difficulty}
\end{figure}

Given the results of 9 models from Table~\ref{tab:pgm-mean-accuracy} that were evaluated in each PGM regime, we compare in Fig.~\ref{fig:pgm-regime-difficulty} the levels of difficulty of individual regimes.
It can be easily noticed that the regimes can be categorised into three distinct groups, depending on their difficulty.
Firstly, there is the Neutral regime, where by definition the test split has the same distribution as the validation split.
In this conventional setup, the performance on validation split is an accurate indicator of the results on the test split.
This, however, is no longer the case in the remaining regimes.
In the Interpolation, Held-out Triple Pairs and Held-out Attribute Pairs regimes, the accuracy of the models noticeably decreases between validation and test splits (by as large as 34.7\% in case of MXGNet $\beta=0$ in Held-out Attribute Pairs regime, see Table~\ref{tab:pgm}).
Generalisation performance practically diminishes in the remaining regimes, i.e.\ Held-out Triples, Extrapolation, Held-out Line-Type and Held-out Shape-Color, where some models even present the behaviour on the test set indistinguishable from random guessing.
These observations demonstrate that although current approaches present satisfactory results in the Neutral regime, all other, more demanding regimes, remain a challenge.
We believe that future works should detour from pushing the limits of i.i.d.\ generalisation and instead explicitly focus on building models capable of generalising to out-of-distribution samples.

\begin{table}[t]
    \centering
    \caption{
    \textbf{RAVEN vs I-RAVEN accuracy.}
    Mean accuracy on the test splits for all configurations of both RAVEN~\cite{zhang2019raven} and I-RAVEN~\cite{hu2021stratified} datasets.
    The models are arranged according to their score on I-RAVEN and then on RAVEN in ascending order.
    The results of the best reported configuration are presented for each model.
    \protect\footnotemark[2]MRNet was evaluated on RAVEN-FAIR.}
    \label{tab:raven-i-raven}
    \begin{tabular}{l|c|c}
        \toprule
        \multirow{2}{*}{Method} & \multicolumn{2}{c}{Test accuracy (\%)} \\
        & RAVEN         & I-RAVEN \\
        \midrule
        CNN LSTM + DRT~\cite{zhang2019raven}                   & 14.0          & -       \\
        WReN + DRT~\cite{zhang2019raven}                       & 15.0          & -       \\
        ARNe~\cite{hahne2019attention}                         & 19.7          & -       \\
        MCPT~\cite{zhuo2020solving}                            & 28.5          & -       \\
        WReN-Tag-Aux~\cite{zhang2019learning}                  & 34.0          & -       \\
        CNN MLP~\cite{zhang2019raven}                          & 37.0          & -       \\
        CNN MLP + DRT~\cite{zhang2019raven}                    & 39.4          & -       \\
        PRD~\cite{kiat2020pairwise}                            & 50.7          & -       \\
        ResNet-18~\cite{zhuo2020solving}                       & 77.2          & -       \\
        LEN + TM~\cite{zheng2019abstract}                      & 78.3          & -       \\
        MXGNet~\cite{wang2020abstract}                         & 83.9          & -       \\
        \textrm{\footnotemark[2]}MRNet~\cite{benny2020scale}                            & 84.0          & -       \\
        ResNet-50 + pre-train~\cite{zhuo2020solving}           & 86.3          & -       \\
        Rel-Base~\cite{spratley2020closer}                     & 91.7          & -       \\
        CoPINet + AL~\cite{kim2020few}                         & 93.5          & -       \\
        DCNet~\cite{zhuo2021effective}                         & 93.6          & -       \\
        CoPINet + ACL~\cite{kim2020few}                        & 93.7          & -       \\
        Rel-AIR~\cite{spratley2020closer}                      & \textbf{94.1} & -       \\
        \midrule
        Context-blind ResNet~\cite{hu2021stratified}           & 71.9          & 12.2    \\
        Context-blind SCL~\cite{wu2020scattering}              & \textbf{94.2} & 12.2    \\
        Context-blind CoPINet~\cite{hu2021stratified}          & \textbf{94.2} & 14.2    \\
        \midrule
        CNN LSTM~\cite{zhang2019raven,hu2021stratified}        & 13.1          & 18.9    \\
        WReN~\cite{zhang2019raven,hu2021stratified}            & 14.7          & 23.8    \\
        ResNet-18~\cite{zhang2019raven,hu2021stratified}       & 53.4          & 40.3    \\
        ResNet-18 + DRT~\cite{zhang2019raven,hu2021stratified} & 59.6          & 40.4    \\
        LEN~\cite{zheng2019abstract,hu2021stratified}          & 72.9          & 41.4    \\
        Wild ResNet~\cite{hu2021stratified}                    & -             & 44.3    \\
        CoPINet~\cite{zhang2019learning,hu2021stratified}      & 91.4          & 46.1    \\
        NCD~\cite{zhuo2021unsupervised}      & 37.0          & 48.2    \\
        CoPINet MLCL+DA~\cite{malkinski2020multilabel}         & -             & 57.1    \\
        SRAN~\cite{hu2021stratified}                           & -             & 60.8    \\
        SRAN MLCL+DA~\cite{malkinski2020multilabel}            & -             & 73.3    \\
        \footnotemark[2]MRNet~\cite{benny2020scale}            & -             & 86.8    \\
        SCL~\cite{wu2020scattering}                            & 91.6          & 95.0    \\
        SCL MLCL+DA~\cite{malkinski2020multilabel}             & -             & \textbf{96.8}    \\
        \midrule
        Human~\cite{zhang2019raven}                            & 84.4          & -       \\
        \bottomrule
    \end{tabular}
\end{table}

\subsection{Results on (I-)RAVEN}
Besides PGM, the methods for solving RPMs are often evaluated on the RAVEN dataset and its derivatives.
The aggregated accuracy scores of the discussed approaches on test splits of both RAVEN and I-RAVEN are shown in Table~\ref{tab:raven-i-raven}.
Firstly, it can be seen that the upper part of the table contains methods that were evaluated only on RAVEN\@.
However, due to the hidden bias in the answer generation algorithm~\cite{hu2021stratified} the results reported on this dataset are inconclusive and can be misleading.
This was demonstrated by remarkable performance of context-blind models, on the one hand, and by significant (45.3 p.p.) drop of accuracy of CoPINet on the balanced dataset (91.4\% on RAVEN vs 46.1\% on I-RAVEN), on the other hand.
Therefore, we strongly advocate for evaluating existing and future DL approaches on the unbiased version of the dataset.

Best results on I-RAVEN are achieved by SCL, a model which also performed very well on PGM\@.
This suggests that SCL is, overall, the best-performing model for solving RPMs.
This claim is further strengthened in the following section which sheds light on the relation between model performance and the number of its parameters.

Detailed performance results on particular RAVEN and I-RAVEN configurations are presented in the \emph{supplementary material}.

\begin{table}[t]
    \centering
    \caption{
    \textbf{Model size.}
    Number of trainable parameters based on model's open-source implementation.
    C-B stands for Context-blind.
    \protect\footnotemark[2]In C-B ResNet-18 we have changed the number of input channels from 16 to 8 in the provided implementation.
    }
    \begin{tabular}{l|N{7}{0}|r}
        \toprule
        Model                                 & {\# params} & GitHub repository                                                                     \\
        \midrule
        SCL~\cite{wu2020scattering}           & 137286      & \href{https://github.com/dhh1995/SCL}{dhh1995/SCL}                                    \\
        CNN LSTM~\cite{zhang2019raven}        & 143960      & \href{https://github.com/WellyZhang/RAVEN}{WellyZhang/RAVEN}                          \\
        CNN MLP~\cite{zhang2019raven}         & 299400      & \href{https://github.com/WellyZhang/RAVEN}{WellyZhang/RAVEN}                          \\
        CNN LSTM + DRT~\cite{zhang2019raven}  & 804628      & \href{https://github.com/WellyZhang/RAVEN}{WellyZhang/RAVEN}                          \\
        WReN~\cite{barrett2018measuring}      & 1216173     & \href{https://github.com/Fen9/WReN}{Fen9/WReN}                                        \\
        Rel-Base~\cite{spratley2020closer}    & 1226673     & \href{https://github.com/SvenShade/Rel-AIR}{SvenShade/Rel-AIR}                        \\
        CoPINet~\cite{zhang2019learning}      & 1685949     & \href{https://github.com/WellyZhang/CoPINet}{WellyZhang/CoPINet}                      \\
        Rel-AIR~\cite{spratley2020closer}     & 1948644     & \href{https://github.com/SvenShade/Rel-AIR}{SvenShade/Rel-AIR}                        \\
        CNN MLP + DRT~\cite{zhang2019raven}   & 2054724     & \href{https://github.com/WellyZhang/RAVEN}{WellyZhang/RAVEN}                          \\
        LEN~\cite{zheng2019abstract}          & 5520673     & \href{https://github.com/zkcys001/distracting_feature}{zkcys001/distracting\_feature} \\
        DCNet~\cite{zhuo2021effective}        & 5833025     & \href{https://github.com/visiontao/dcnet}{visiontao/dcnet}                            \\
        NCD~\cite{zhuo2021unsupervised}       & 11177025    & \href{https://github.com/visiontao/ncd}{visiontao/ncd}                          \\
        \footnotemark[2]C-B ResNet-18~\cite{zhang2019raven}   & 11474342    & \href{https://github.com/WellyZhang/RAVEN}{WellyZhang/RAVEN}                          \\
        ResNet-18~\cite{zhang2019raven}       & 11499430    & \href{https://github.com/WellyZhang/RAVEN}{WellyZhang/RAVEN}                          \\
        ResNet-18 + DRT~\cite{zhang2019raven} & 13254754    & \href{https://github.com/WellyZhang/RAVEN}{WellyZhang/RAVEN}                          \\
        MRNet~\cite{benny2020scale}           & 19531841    & \href{https://github.com/yanivbenny/MRNet}{yanivbenny/MRNet}                          \\
        SRAN~\cite{hu2021stratified}          & 44030217    & \href{https://github.com/husheng12345/SRAN}{husheng12345/SRAN}                        \\
        \bottomrule
    \end{tabular}
    \label{tab:num-params}
\end{table}

\begin{figure*}[t]
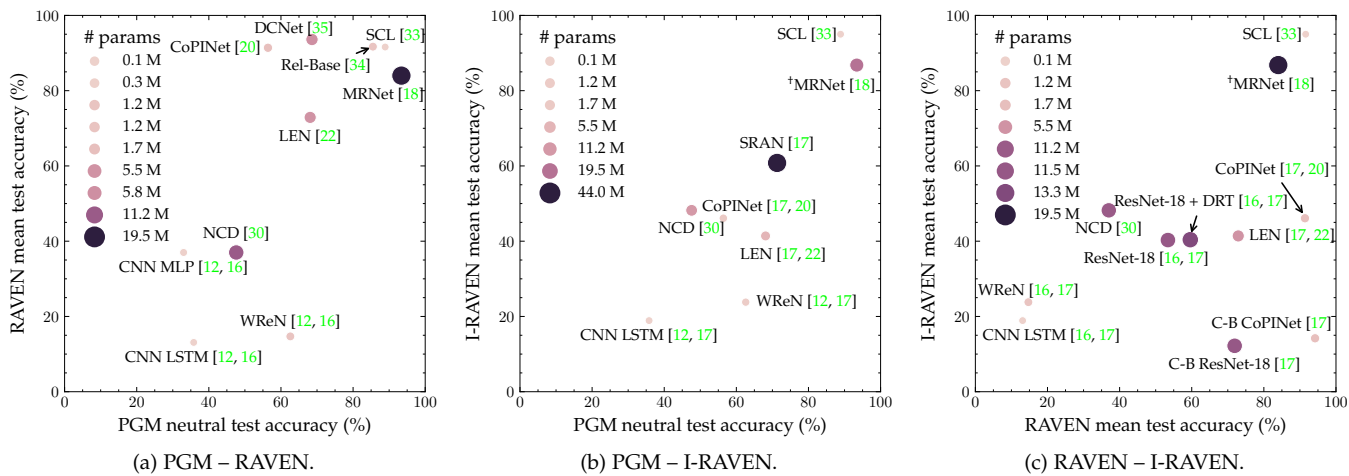

    \centering
    \subfloat[PGM -- RAVEN.]{
        \resizebox{.32\textwidth}{!}{\input{images/results/num_params_pgm_raven.pgf}}
    }
    \hfil
    \subfloat[PGM -- I-RAVEN.]{
        \resizebox{.32\textwidth}{!}{\input{images/results/num_params_pgm_i-raven.pgf}}
    }
    \hfil
    \subfloat[RAVEN -- I-RAVEN.]{
        \resizebox{.32\textwidth}{!}{\input{images/results/num_params_raven_i-raven.pgf}}
    }
    \caption{
    \textbf{Size vs accuracy.}
    Model performance evaluated on PGM, RAVEN and I-RAVEN against the number of its trainable parameters.
    The plots present only those models which were evaluated in the literature on the respective pairs of datasets.
    C-B stands for Context-blind.
    \protect\footnotemark[2]MRNet was evaluated on RAVEN-FAIR instead of I-RAVEN.
    }
    \label{fig:num-params}
\end{figure*}

\subsection{Model sizes}

To the best of our knowledge, in the existing literature on DL methods for solving RPMs, the size of the models---measured in the number of parameters---was never discussed.
However, some models, e.g.\ MRNet~\cite{benny2020scale} or SRAN~\cite{hu2021stratified} use deep visual backbones based on ResNet~\cite{he2016deep} which results in huge number of trainable parameters.
The numbers of parameters of all discussed models with published open-source implementations are compared in Table~\ref{tab:num-params}.
Model sizes range from hundreds of thousands up to tens of millions, with the largest one (SRAN) having around 320 times more parameters than the smallest one (SCL).

Although a large number of trainable parameters is often necessary in various real-world problems, e.g.\ in computer vision~\cite{goyal2021self,kolesnikov2020big} or natural language processing~\cite{radford2019language,brown2020language} where practically unlimited supply of the training data exists, in the context of RPMs this is no longer the case.
Figure~\ref{fig:num-params} shows that in RPM benchmarks the best results are achieved by the model with the smallest number of parameters -- SCL\@.
Although a bit surprising, this observation shows that a relatively small model can effectively learn to solve RPMs with competitive generalisation ability (cf.\ Table~\ref{tab:pgm-mean-accuracy}).
In other words, these results prove that in the case of RPMs, large (in terms of the number of parameters) visual backbones (e.g. deep ResNets) are not necessary, and instead, the existence of efficient parameter-sharing modules is crucial.

While this is yet to be verified in other AVR tasks, we hypothesize that due to their visual similarity (e.g. lack of texture and presence of 2D greyscale shapes), analogous architectural decisions should be made when designing AVR machine solvers in general.

    \section{Discussion}\label{sec:discussion}

The initial reason for introducing RPMs into DL literature, was to use these problems as a proxy for estimating machine intelligence.
It turned out, however, that the ability of spatial and abstract reasoning is also crucial for the development of intelligent systems in various other settings~\cite{zhu2020dark}.
Consequently, methods proposed in the context of RPMs are oftentimes relevant in other research and practical contexts.
In this section, we link the discussed approaches to advancements in other fields and highlight the main unsolved challenges and open questions left for investigation in future work.

\subsection{Seeds and fruits of RPM research}
Relation Network, the fundamental component of discussed models, e.g. WReN, LEN or MLRN, has already demonstrated its usefulness in multiple tasks.
In~\cite{park20183d} the authors consider a 3D human pose estimation problem~\cite{sarafianos20163d,chen2020monocular,zheng2020deep} and propose a DL algorithm to tackle this challenge.
The designed method employs an RN to capture relations among different body parts.
Similarly, in the context of semantic segmentation~\cite{garcia2018survey,hao2020brief,minaee2021image}, the idea of capturing long-distance spatial relationships between entities using RN is further explored.
It is shown that a relational reasoning component can be used to augment CNN feature maps by exploiting both channel-wise and spatial feature relations~\cite{mou2019relation}.
Such incorporation of the global context is a recurring theme in works that exploit RNs (and is also explored in other neural modules~\cite{battaglia2016interaction,wang2018non,hu2018relation,chen2019graph}).
RN was further applied to model spatio-temporal interactions between human actors, objects and scene elements in the context of action recognition~\cite{sun2018actor}, to train an effective image recognition model in a contrastive self-supervised setting~\cite{patacchiola2020self}, or to provide a mechanism for relational reasoning over structured representations for a deep reinforcement learning agent~\cite{zambaldi2018deep}.

In order to successfully solve RPMs and learn to formulate analogies, multiple discussed works have employed various forms of contrastive mechanisms either directly in the model architecture~\cite{zhang2019learning,zhuo2021effective} or in the objective function~\cite{zhang2019learning,malkinski2020multilabel,kim2020few}.
Contrastive approaches~\cite{gutmann2010noise,hyvarinen2016unsupervised,oord2018representation} are especially useful for self-supervised learning, where the availability of labelled data is scarce.
The importance of such methods was already demonstrated in the context of computer vision~\cite{chen2020simple,he2020momentum,patacchiola2020self}, natural language processing~\cite{mikolov2013distributed,saunshi2019theoretical,klein2020contrastive}, speech recognition~\cite{schneider2019wav2vec,kreuk2020self,al2021clar}, or reinforcement learning~\cite{kipf2020contrastive,laskin2020curl,liu2021returnbased}.

RPM solving methods are often directly applicable, after minor adjustments only, to related abstract reasoning tasks.
RN was found to be competitive to other models in solving arithmetic visual reasoning tasks~\cite{zhang2020machine}, while WReN-based models were found to be one of the top performing methods in solving the visual analogy problems~\cite{hill2019learning} and abstract reasoning matrices structurally similar to RPMs~\cite{van2019disentangled}.

\begin{figure}[t]
    \centering
    \subfloat[RAVEN, $i\in\{0, 1, \ldots, 6\}$]{
        \includegraphics[width=.23\textwidth]{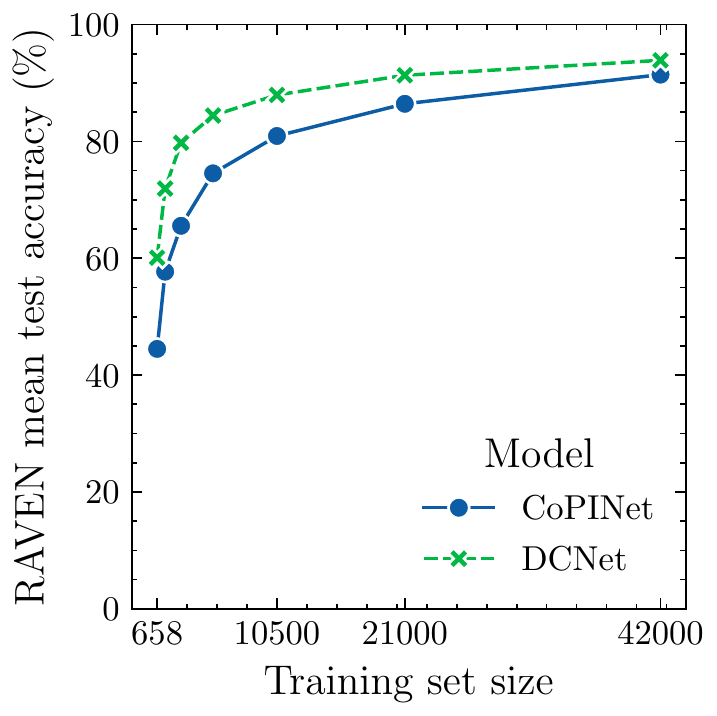}
    }
    \hfil
    \subfloat[PGM, $i \in \{0, 2, \ldots, 12\}$]{
        \includegraphics[width=.23\textwidth]{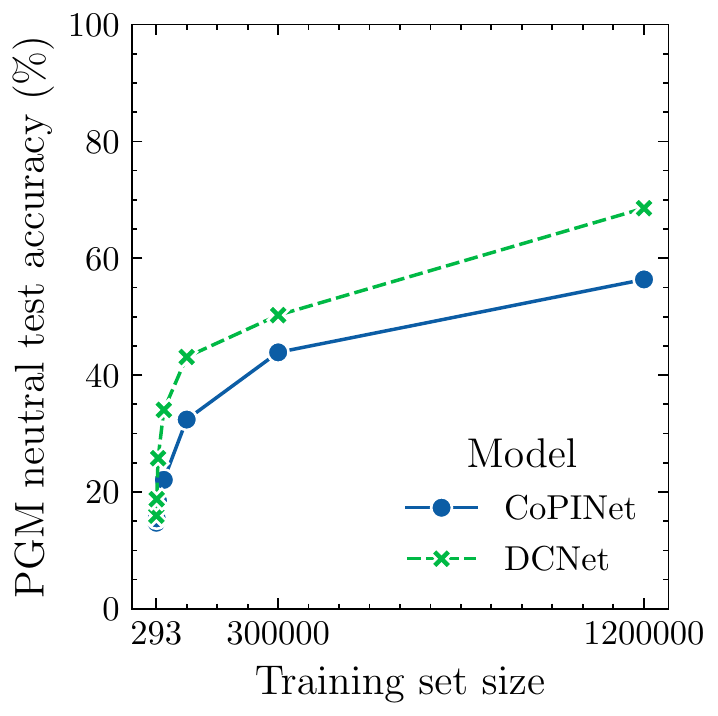}
    }
    \caption{
    \textbf{Low-sample regime.}
    The plots present the performance of two models trained on splits of a) RAVEN and b) PGM datasets.
    The split sizes equal to $N/2^i$, where $N$ denotes the size of the original dataset, which equals to \numprint{42000} for RAVEN and \numprint{1200000} for PGM.
    The results were reported in~\cite{zhuo2021effective}.
    }
    \label{fig:rpm-low-sample}
\end{figure}

\subsection{Challenges and open problems}
Even though the discussed works have embarked on a quest to measure machine intelligence by means of evaluating their performance on RPM benchmarks, some may oppose the validity of this path.
When solving an RPM, a human solver is often faced with a task that he/she has not encountered before, which tests the ability of adaptive problem solving.
Contrary to humans, the majority of current DL approaches use thousands~\cite{zhang2019raven} or millions~\cite{barrett2018measuring} training samples beforehand.
In addition, the performance of these models rapidly deteriorates when the size of the training corpora decreases, as illustrated in Fig.~\ref{fig:rpm-low-sample}.
Conversely, humans are able not only to grasp the concepts after familiarising themselves with just a few examples, but also to extrapolate knowledge gained when solving simple matrices to more advanced ones.
Given this notable contrast it is worth to advocate the search for efficient methods for solving RPMs without access to huge training sets.
We are convinced that it is worthwhile to pursue certain emerging pathways discussed in this paper that are explicitly designed to probe DL algorithms in few-shot learning setups~\cite{chollet2019measure,nie2020bongard}.

Another way of ensuring that the developed pattern analysis algorithms are benchmarked similarly to humans in new environments is to follow the perspective of the seminal PGM paper~\cite{barrett2018measuring}.
By explicitly defining various generalisation regimes ~\citet{barrett2018measuring} allowed to directly measure generalisation (performance in new settings) of DL methods for solving RPMs.
However, as of today no effective method capable of achieving human-like performance in all regimes was proposed.
In fact, existing approaches seem to possess specific generalisation abilities that are rather a side effect than a deliberate choice.
While recent work~\cite{webb2020learning} shows that neural models can generalise in AVR tasks that focus on extrapolation, which constitutes one of the most demanding PGM regimes, there is still a long way to construct a universal learning system that would excel in all regimes.

Another key characteristic that differentiates humans from current AI systems is the ability to solve various types of AVR problems with limited training.
While humans are known to be able to generalise and transfer knowledge between problems, such property has only been briefly demonstrated by the existing DL methods.
Some works that evaluate DL approaches on the RAVEN dataset show that models trained on one configuration may learn to build abstractions useful for solving matrices belonging to other configurations~\cite{zhang2019raven,spratley2020closer,zhang2021abstract}.
In~\cite{mandziuk2019deepiq}, an RPM solving model was shown to be adaptable to solving odd-one-out tasks that consisted of similar input images.
While the above examples indicate, to some extent, the ability of DL algorithms to i.i.d.\ generalisation in the context of AVR, knowledge reuse between multiple distinct AVR problems remains to be investigated in future work.

    \section{Conclusion}\label{sec:conclusion}

This paper discusses recent progress in applying DL methods to solving RPMs, summarises various methods of learning to solve these tasks, reviews the existing RPM benchmark sets, and categorises the DL models employed in this field.
Also, by aggregating results of recently published methods, it brings to attention the most challenging aspects of RPM problems, which to this day remain primarily unsolved.

The paper argues that while RPMs were initially proposed as a task for measuring human intelligence and were later employed as a proxy for estimating machine intelligence, they additionally offer a comprehensible playground for developing and testing abstract and relational reasoning approaches.
Viewed from this perspective, advancements in RPM research are applicable to a broad spectrum of other domains where spatial and abstract reasoning skills are required.

Even though the task of solving RPMs has seen a wide interest within the DL community in recent years, its core challenges remain unattained.
We hope that by collating the advances in methods for solving RPMs, this survey will stimulate progress in future AVR research.

    \bibliographystyle{IEEEtranN}
    \bibliography{main}

    \vfill
    \eject
    \onecolumn
    \appendix[Detailed results]

Additional results on particular configurations from RAVEN~\cite{zhang2019raven} and I-RAVEN~\cite{hu2021stratified} datasets are provided in Table~\ref{tab:raven} and Table~\ref{tab:i-raven}, respectively.

\begin{table*}[!h]
    \centering
    \caption{
    \textbf{RAVEN accuracy.}
    Accuracy on the test split of the RAVEN dataset~\cite{zhang2019raven}.
    Mean denotes the mean accuracy for all configurations.
    The \texttt{Left-Right} configuration is denoted as \texttt{L-R}, \texttt{Up-Down} as \texttt{U-D}, \texttt{Out-InCenter} as \texttt{O-IC} and \texttt{Out-InGrid} as \texttt{O-IG}.
    }
    \begin{tabular}{l|c|ccccccc}
        \toprule
        \multirow{2}{*}{Method} & \multicolumn{8}{c}{Test accuracy (\%)} \\
        & Mean          & \texttt{Center} & \texttt{2x2Grid} & \texttt{3x3Grid} & \texttt{L-R} & \texttt{U-D} & \texttt{O-IC} & \texttt{O-IG} \\
        \midrule
        CNN LSTM~\cite{zhang2019raven}                & 13.1          & 13.2            & 14.1             & 13.7             & 12.8         & 12.4         & 12.2          & 13.0          \\
        CNN LSTM + DRT~\cite{zhang2019raven}          & 14.0          & 14.3            & 15.1             & 14.1             & 13.8         & 13.2         & 14.0          & 13.3          \\
        WReN~\cite{zhang2019raven}                    & 14.7          & 13.1            & 28.6             & 28.3             & 7.5          & 6.3          & 8.4           & 10.6          \\
        WReN + DRT~\cite{zhang2019raven}              & 15.0          & 15.4            & 23.3             & 29.5             & 7.0          & 8.4          & 8.9           & 12.4          \\
        ARNe~\cite{hahne2019attention}                & 19.7          & -               & -                & -                & -            & -            & -             & -             \\
        MCPT~\cite{zhuo2020solving}                   & 28.5          & 35.9            & 26.0             & 27.2             & 29.3         & 27.4         & 33.1          & 20.7          \\
        WReN-Tag-Aux~\cite{zhang2019learning}         & 34.0          & 58.4            & 38.9             & 37.7             & 21.6         & 19.7         & 38.8          & 22.6          \\
        NCD~\cite{zhuo2021unsupervised} & 37.0 & 45.5 & 35.5 & 39.5 & 34.9 & 33.4 & 40.3 & 30.0 \\
        CNN MLP~\cite{zhang2019raven}                 & 37.0          & 33.6            & 30.3             & 33.5             & 39.4         & 41.3         & 43.2          & 37.5          \\
        CNN MLP + DRT~\cite{zhang2019raven}           & 39.4          & 37.3            & 30.1             & 34.6             & 45.5         & 45.5         & 45.9          & 37.5          \\
        PRD~\cite{kiat2020pairwise}                   & 50.7          & 74.6            & 38.7             & 34.9             & 60.8         & 60.3         & 62.5          & 23.4          \\
        ResNet-18~\cite{zhang2019raven}               & 53.4          & 52.8            & 41.9             & 44.3             & 58.8         & 60.2         & 63.2          & 53.1          \\
        ResNet-18 + DRT~\cite{zhang2019raven}         & 59.6          & 58.1            & 46.5             & 50.4             & 65.8         & 67.1         & 69.1          & 60.1          \\
        LEN~\cite{zheng2019abstract}                  & 72.9          & 80.2            & 57.5             & 62.1             & 73.5         & 81.2         & 84.4          & 71.5          \\
        ResNet-18~\cite{zhuo2020solving}              & 77.2          & 72.8            & 57.0             & 62.7             & 91.0         & 89.6         & 88.4          & 78.9          \\
        LEN + TM~\cite{zheng2019abstract}             & 78.3          & 82.3            & 58.5             & 64.3             & 87.0         & 85.5         & 88.9          & 81.9          \\
        MXGNet~\cite{wang2020abstract}                & 83.9          & -               & -                & -                & -            & -            & -             & -             \\
        MRNet~\cite{benny2020scale}                   & 84.0          & -               & -                & -                & -            & -            & -             & -             \\
        ResNet-50 + pre-train~\cite{zhuo2020solving}  & 86.3          & 89.5            & 66.6             & 68.0             & 97.9         & 98.2         & 96.6          & 87.2          \\
        CoPINet~\cite{zhang2019learning}              & 91.4          & 95.1            & 77.5             & 78.9             & 99.1         & 99.7         & 98.5          & 91.4          \\
        SCL~\cite{wu2020scattering}                   & 91.6          & 98.1            & 91.0             & 82.5             & 96.8         & 96.5         & 96.0          & 80.1          \\
        Rel-Base~\cite{spratley2020closer}            & 91.7          & 97.6            & 85.9             & 86.9             & 93.5         & 96.5         & 97.6          & 83.8          \\
        CoPINet + AL~\cite{kim2020few}                & 93.5          & 98.6            & 80.5             & 83.2             & 99.7         & 99.8         & 99.4          & 93.3          \\
        DCNet~\cite{zhuo2021effective}                & 93.6          & 97.8            & 81.7             & 86.7             & 99.8         & 99.8         & 99.0          & 91.5          \\
        CoPINet + ACL~\cite{kim2020few}               & 93.7 & 98.4            & 81.0             & 84.0             & 99.7         & 99.8         & 99.4          & 93.9          \\
        Rel-AIR~\cite{spratley2020closer}             & \textbf{94.1}          & 99.0            & 92.4             & 87.1             & 98.7         & 97.9         & 98.0          & 85.3          \\
        \midrule
        Context-blind ResNet~\cite{hu2021stratified}  & 71.9          & -               & -                & -                & -            & -            & -             & -             \\
        Context-blind SCL~\cite{wu2020scattering} & \textbf{94.2} & -               & -                & -                & -            & -            & -             & -             \\
        Context-blind CoPINet~\cite{hu2021stratified} & \textbf{94.2} & -               & -                & -                & -            & -            & -             & -             \\
        \midrule
        Human~\cite{zhang2019raven}                   & 84.4          & 95.4            & 81.8             & 79.6             & 86.4         & 81.8         & 86.4          & 81.8          \\
        \bottomrule
    \end{tabular}
    \label{tab:raven}
\end{table*}

\begin{table*}[!h]
    \centering
    \caption{
    \textbf{I-RAVEN accuracy.}
    Accuracy on the test split of the I-RAVEN dataset~\cite{hu2021stratified}.
    Mean denotes the mean accuracy for all configurations.
    The \texttt{Left-Right} configuration is denoted as \texttt{L-R}, \texttt{Up-Down} as \texttt{U-D}, \texttt{Out-InCenter} as \texttt{O-IC} and \texttt{Out-InGrid} as \texttt{O-IG}.
    \protect\footnotemark[2]MRNet was evaluated on RAVEN-FAIR in~\cite{benny2020scale}.
    }
    \begin{tabular}{l|c|ccccccc}
        \toprule
        \multirow{2}{*}{Method} & \multicolumn{8}{c}{Test accuracy (\%)} \\
        & Mean          & \texttt{Center} & \texttt{2x2Grid} & \texttt{3x3Grid} & \texttt{L-R} & \texttt{U-D} & \texttt{O-IC} & \texttt{O-IG} \\
        \midrule
        CNN LSTM~\cite{hu2021stratified}               & 18.9          & 26.2            & 16.7             & 15.1             & 14.6         & 16.5         & 21.9          & 21.1          \\
        WReN~\cite{hu2021stratified}                   & 23.8          & 29.4            & 26.8             & 23.5             & 21.9         & 21.4         & 22.5          & 21.5          \\
        ResNet-18~\cite{hu2021stratified}              & 40.3          & 44.7            & 29.3             & 27.9             & 51.2         & 47.4         & 46.2          & 35.8          \\
        ResNet-18 + DRT~\cite{hu2021stratified}        & 40.4          & 46.5            & 28.8             & 27.3             & 50.1         & 49.8         & 46.0          & 34.2          \\
        LEN~\cite{hu2021stratified}                    & 41.4          & 56.4            & 31.7             & 29.7             & 44.2         & 44.2         & 52.1          & 31.7          \\
        Wild ResNet~\cite{hu2021stratified}            & 44.3          & 50.9            & 33.1             & 30.8             & 53.1         & 52.6         & 50.9          & 38.7          \\
        CoPINet~\cite{hu2021stratified}                & 46.1          & 54.4            & 36.8             & 31.9             & 51.9         & 52.5         & 52.2          & 42.8          \\
        NCD~\cite{zhuo2021unsupervised} & 48.2 & 60.0 & 31.2 & 30.0 & 58.9 & 57.2 & 62.4 & 39.0 \\
        CoPINet MLCL+DA~\cite{malkinski2020multilabel} & 57.1          & -               & -                & -                & -            & -            & -             & -             \\
        SRAN~\cite{hu2021stratified}                   & 60.8          & 78.2            & 50.1             & 42.4             & 70.1         & 70.3         & 68.2          & 46.3          \\
        SRAN MLCL+DA~\cite{malkinski2020multilabel}    & 73.3          & -               & -                & -                & -            & -            & -             & -             \\
        \footnotemark[2]MRNet~\cite{benny2020scale}    & 86.8          & 97.0            & 72.7             & 69.5             & 98.7         & 98.9         & 97.6          & 73.3          \\
        SCL~\cite{wu2020scattering}                    & 95.0          & 99.0            & 96.2             & 89.5             & 97.9         & 97.1         & 97.6          & 87.7          \\
        SCL MLCL+DA~\cite{malkinski2020multilabel}     & \textbf{96.8} & -               & -                & -                & -            & -            & -             & -             \\
        \midrule
        Context-blind SCL~\cite{wu2020scattering}      & 12.2          & -               & -                & -                & -            & -            & -             & -             \\
        Context-blind ResNet~\cite{hu2021stratified}   & 12.2          & -               & -                & -                & -            & -            & -             & -             \\
        Context-blind CoPINet~\cite{hu2021stratified}  & 14.2          & -               & -                & -                & -            & -            & -             & -             \\
        \bottomrule
    \end{tabular}
    \label{tab:i-raven}
\end{table*}

\end{document}